\documentclass{article}

\usepackage{microtype}
\usepackage{graphicx}
\usepackage{subfigure}
\usepackage{booktabs} 

\usepackage{appendix}
\usepackage{selectp}

\usepackage{amsmath,amsfonts,bm}

\def\eqref#1{equation~\ref{#1}}

\def\1{\bm{1}}

\DeclareMathAlphabet{\mathsfit}{\encodingdefault}{\sfdefault}{m}{sl}
\SetMathAlphabet{\mathsfit}{bold}{\encodingdefault}{\sfdefault}{bx}{n}

\usepackage[ruled,vlined]{algorithm2e}
\SetKwInput{kwData}{Data}
\SetKwInput{kwInit}{Init}
\SetKwInput{kwInput}{Input}

\usepackage[T1]{fontenc}

\usepackage{hyperref}

\usepackage{color, colortbl}

\usepackage{siunitx}
\usepackage[accepted]{icml2021}

\icmltitlerunning{Imitation by Predicting Observations}

\begin{document}

\newcommand{\mycomment}[3]{{\textcolor{#3}{#1 #2}}}
\newcommand{\drewmarker}[1]{\textcolor{blue}{\emph{Drew:} #1}}
\newcommand{\tododrewmarker}[1]{\textcolor{blue}{\emph{TODO(Drew):} #1}}

\newcommand{\drew}[1]{\mycomment{\drewmarker}{#1}{blue}}
\newcommand{\tododrew}[1]{\mycomment{\tododrewmarker}{#1}{blue}}

\newcommand{\greg}[1]{\textcolor{red}{#1}}
\newcommand{\rob}[1]{\textcolor{magenta}{#1}}
\newcommand{\yury}[1]{\textcolor{green}{#1}}
\newcommand{\arun}[1]{\textcolor{purple}{#1}}

\twocolumn[
\icmltitle{Imitation by Predicting Observations}

\begin{icmlauthorlist}
\icmlauthor{Andrew Jaegle}{dm}
\icmlauthor{Yury Sulsky}{dm}
\icmlauthor{Arun Ahuja}{dm}
\icmlauthor{Jake Bruce}{dm}
\icmlauthor{Rob Fergus}{dm}
\icmlauthor{Greg Wayne}{dm}
\end{icmlauthorlist}

\icmlaffiliation{dm}{DeepMind}

\icmlcorrespondingauthor{Andrew Jaegle}{drewjaegle@deepmind.com}

\icmlkeywords{Imitation Learning, Inverse Reinforcement Learning, Learning from Demonstrations, Imitation from Observations, Predictive Modeling, Maximum entropy inverse reinforcement learning, Model-based, FORM, GAIL}

\vskip 0.3in
]

\printAffiliationsAndNotice{}  

\begin{abstract}

Imitation learning enables agents to reuse and adapt the hard-won expertise of others, offering a solution to several key challenges in learning behavior. Although it is easy to observe behavior in the real-world, the underlying actions may not be accessible. 
We present a new method for imitation solely from observations that achieves comparable performance to experts on challenging continuous control tasks while also exhibiting robustness in the presence of observations unrelated to the task. Our method, which we call FORM (for ``Future Observation Reward Model'') is derived from an inverse RL objective and imitates using a model of expert behavior learned by generative modelling of the expert's observations, without needing ground truth actions. We show that FORM performs comparably to a strong baseline IRL method (GAIL) on the DeepMind Control Suite benchmark, while outperforming GAIL in the presence of task-irrelevant features.

\end{abstract}

\section{Introduction}
\label{introduction}

The goal of imitation is to learn to produce behavior that matches that of an expert on unseen data, given demonstrations of the expert's behavior \cite{ng2004apprenticeship, osa2018algorithmic}. The field of imitation learning offers tools for learning behavior when programmed rewards cannot be provided, or when rewards can only be partially or sparsely specified. Imitation learning has been at the heart of several breakthroughs in building AI agents \cite{pomerleau1989alvinn, abbeel2010autonomous, silver2016mastering, vinyals2019grandmaster, openai2019dota}, allowing agents to learn even when faced with hard exploration problems \cite{gulcehre2020making}.

There is widespread evidence that imitation (among other forms of social learning) is a core mechanism by which humans and other animals learn to acquire a sophisticated behavioral repertoire \cite{tomasello1996doapes, laland2008animal, laland2009animal, huber2009theevolution}. While most algorithms for imitation learning assume that demonstrations contain the actions the expert executed, animals must imitate without directly observing what actions the expert took (i.e. without knowing exactly what commands were issued to produce the observable changes). In the context of machine learning, solving the problem of \textit{imitation from observation} is a key step towards the tantalizing possibility of learning behavior from unlabeled and easy to collect data, such as raw video footage of human activity. Many recent algorithms for imitation have focused on addressing the problem of imitation in very small data regimes, but the challenge in imitating from these abundant sources of data is not primarily one of quantity. The challenge is rather how to learn models for imitation that are general enough to learn and generalize from data that depicts a rich (and unknown) reward structure. In this work, we show how predictive generative models can be used to learn a general reward model from observations alone.

\definecolor{orange}{rgb}{0.8, 0.33, 0.0}
\definecolor{darkgreen}{rgb}{0.2, 0.6, 0.2}
\begin{figure*}[t!]
    \centering
    \includegraphics[width=.33\linewidth]{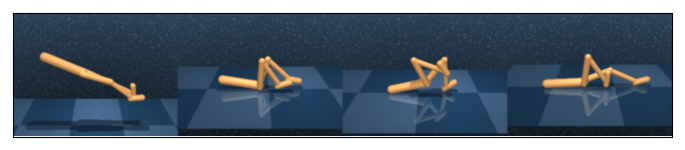}\hfill\
    \includegraphics[width=.33\linewidth]{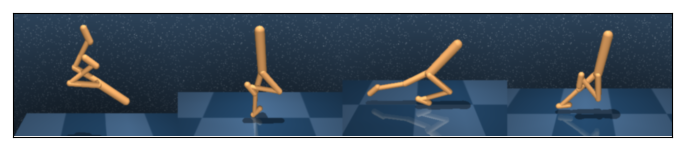}\hfill\
    \includegraphics[width=.33\linewidth]{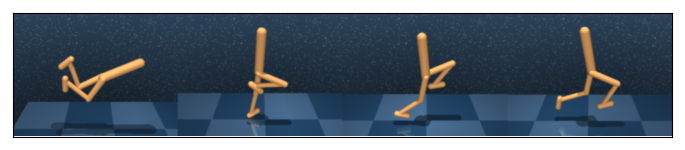}
    
    \includegraphics[width=.33\linewidth]{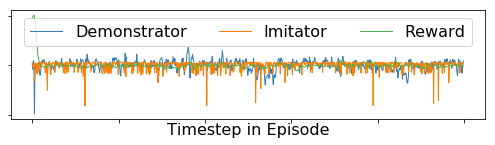}\hfill\
    \includegraphics[width=.33\linewidth]{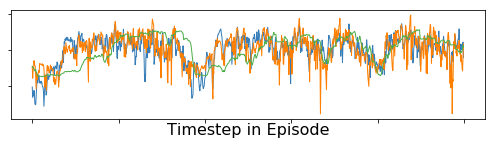}\hfill\
    \includegraphics[width=.33\linewidth]{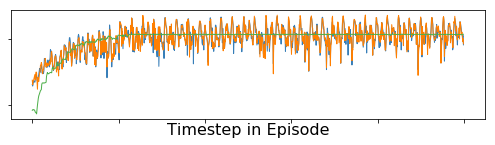}
    \vspace{-7mm}
    \caption{FORM learns to imitate expert behavior using sequences of internal state observations, without access to the expert's actions. Visualizations of agent behavior (top) and reward curves for a single episode (bottom) are shown after 0 (left), 50k (middle) and 5M update steps. FORM imitates using two learned models: both the \textcolor{blue}{demonstrator} model (trained offline) and the \textcolor{orange}{imitator} model (trained online) log-likelihoods track the unseen \textcolor{darkgreen}{task reward} as the imitation agent learns. Agent behavior is shown as images, but we use lower-dimensional internal state observations in this work.}
      \vspace{-4mm}
    \label{fig:trajectory_visualization}
\end{figure*}

Current state-of-the-art approaches to imitation (including from observation) pose learning as an adversarial game: a classifier estimates the probability that a state is visited by the expert or imitator, and the policy seeks to maximize the classifier error \cite{merel2017learning, torabi2019generative}. Because these methods are based on matching the expert's occupancy using a fixed dataset of demonstrations, they tend to be very sensitive to the precise details of the demonstrations and to the representation used. This property makes learning with adversarial methods difficult when using raw, noisy observations without extensive tuning and careful use of strong forms of regularization \cite{peng2019variational}, domain or task knowledge \cite{zolna2020taskrelevant}, or a combination of behavioral cloning and careful representation design \cite{abramson2020imitating}.

In this work, we introduce the future observation reward model (FORM) (see Figure \ref{fig:trajectory_visualization}), which address the problem of imitation from observation while exhibiting both (1) generality and expressiveness by coupling predictive generative models with inverse RL (IRL) and (2) improved robustness by foregoing an adversarial formulation. In FORM, the imitator tries to match the probability of observation sequences in the expert data. It does so using a learned generative model of expert observation sequences and a learned generative model of its own observation sequences. In other words, FORM casts the problem of learning from demonstrations as a sequence prediction problem, using a generative model of expert sequences to guide RL. Because FORM builds separate models of expert and imitator sequences, rather than using a single classifier to discriminate expert and imitator states, it is less prone to focus on irrelevant differences between the expert and imitator demonstrations. The structure of the FORM objective makes it theoretically straightforward to optimize using standard policy optimization tools and, as we show, empirically competitive on the DeepMind Control Suite continuous control benchmark domain.

This stands in contrast to adversarial methods, such as Generative Adversarial Imitation Learning (GAIL), whose objectives are known to be ill-posed (without additional regularization) and challenging to optimize both in theory and in practice \cite{arjovsky2017wasserstein, gulrajani2017improved, mescheder2017numerics}. This property makes it difficult to apply adversarial techniques to imitation in settings with even small differences between expert and imitator settings \cite{zolna2020taskrelevant}. Robustness to distractors is an important part of behavior learning, as recently illustrated by \citealt{Stone21} in the context of RL with image background distractors. These situations are common in practice: the lab environment where expert data is collected for a robot will be quite different to where it might be deployed. While it may be possible to collect a large number of demonstrations, it is impossible to exhaustively sample all possible sources of differences between the two domains (such as the surface texture, robot physical parameters, or environment appearance). These differences confound the signal that must be imitated, leading to the risk of spurious dependencies between the two being learned. As we will show, FORM exhibits greater robustness than a well-tuned adversarial imitation method, GAIL from Observation, or GAIfO \cite{torabi2019generative} in presence of task-independent features.

We make the following technical contributions in this work:
\begin{enumerate}
\vspace{-3mm}
    \item We derive the FORM reward from an objective for inverse reinforcement learning from observations. We show that this reward can be maximized using generative models of expert and imitator behavior with standard policy optimization techniques.
    \item We develop a practical algorithm for imitation learning using the FORM reward and demonstrate that it performs competitively with a well-tuned GAIfO model on the DeepMind Control Suite benchmark.
    \item We show that FORM is more robust than GAIfO in the presence of extraneous, task-irrelevant features, which simulate domain shift between expert and imitator settings.
\end{enumerate}

\section{Background and related work}
\label{related_work}

\textbf{RL, IRL, and imitation} Reinforcement learning is concerned with learning a policy that maximizes the expected return, which is given as the expected sum of all future discounted rewards \cite{sutton2018reinforcement}, which are typically observed. In imitation learning, on the other hand, we are not given a reward function, but we do have access to demonstrations produced by a demonstrator (or expert) policy, which maximizes some (unobserved) expected return. 

IRL has the related goal of recovering the unobserved reward function from expert behavior. IRL offers a general formula for imitation: estimate the reward function underlying the demonstration data (a ``reward model'') and maximize this reward by RL \cite{ng2000algorithms}, possibly iterating multiple times until convergence. Alternative approaches to imitation, such as behavioral cloning (BC) \cite{pomerleau1989alvinn} or BC from observations (BCO) \cite{torabi2018behavioral}, typically have difficulty producing reliable behavior away from configurations seen in the expert demonstrations. This is because small errors in predicting actions or mimicking short-term agent behavior accumulates over long behavioral timescales.\footnote{The standard solution to this problem for BC assumes access to an expert policy that can be repeatedly queried \cite{ross2011reduction}, which is not always feasible.} IRL methods like FORM avoid this problem: because they perform RL on a learned reward, they can learn through experience to recover from mistakes by focusing on long-term consequences of each action.

\textbf{GAIL and occupancy-based IRL} Most contemporary IRL-based approaches to imitation -- as exemplified by GAIL -- use a strategy of state-action occupancy matching, typically by casting imitation as an adversarial game and learning a classifier to discriminate states and actions sampled uniformly from the expert demonstrations from those encountered by the imitator \cite{ho2016generative, torabi2019generative, fu2018learning, kostrikov2019discriminator, ghasemipour2019divergence}. In contrast, rather than classifying states as belonging to the expert or imitator, FORM learns to imitate using separate generative models of expert and imitator behavior. This means that FORM is built on predictive models of the form $p(x_t|x_{t-1})$, where $x$s are observations, rather than a single model of the form $p(\text{expert} | x)$ that tries to classify observations as generated by the expert or not. FORM's objective is similar in spirit to classical feature-matching and maximum-entropy formulations of imitation \cite{ng2000algorithms, ng2004apprenticeship, ziebart2008maximum}, while also providing a fully probabilistic interpretation and making minimal assumptions about the environment (the FORM objective does not require an MDP or deterministic transitions).

\begin{figure}[t]
    \centering
    \includegraphics[keepaspectratio,width=.35\linewidth]{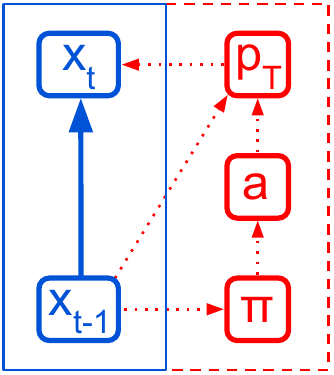}
    \vspace{-8pt}
    \caption{FORM's demonstrator and imitator effect models are \textit{effect models}, generative models $p(x_{t} | x_{t-1})$ of the change in observation (\textcolor{blue}{observed}) produced by a policy $\pi$ in an environment with transition dynamics $p_T$ (\textcolor{red}{unobserved}). The models used in model-based RL are usually of the form $p(x_t | x_{t-1}, a_{t-1})$ and aim to model transition dynamics rather than the full distribution of outcomes given a policy.}
    \label{fig:effect_model}
    \vspace{-12pt}
\end{figure}

\textbf{Other related methods for imitation} Other recent work has used generative models in the context of imitation learning: this work typically retains GAIL's occupancy-based perspective \cite{baram2016model, jarrett2020strictly, liu2021energy} or introduces a generative model to provide a heuristic reward \cite{yu2020intrinsic}. Unlike FORM, which uses effect models (see Figure \ref{fig:effect_model}) that are suitable for imitation from observations, this work models quantities that are useful primarily in conjunction with actions (modeling state-action densities and/or dynamics models for GAIL augmentation). Other recently proposed methods learn reward models either purely or partially offline \cite{kostrikov2020imitation, jarrett2020strictly, arenz2020nonadversarial}. This approach leans on the presence of actions in the demonstrator data. Although FORM's demonstrator effect model is learned offline, FORM's online phase is essential to the process of distilling an effect model (which doesn't use actions) into a policy (which does).

\textbf{Learning to act from observations} Many methods have been proposed for imitation from observation \cite{torabi2019recent}, but most methods that do so using IRL are based around GAIL \cite{wang2017robust, torabi2019generative, sun2019provably}. Recent work has obtained interesting results using solutions based around tracking or matching trajectories in learned feature spaces \cite{peng2018deepmimic, merel2019neural}, by matching imitator actions to learned models of expert trajectories (interpreted as inverse models of the expert action) \cite{schmeckpeper2020learning, zhu2020offpolicy, edwards2019imitating, pathak2018zeroshot}, and by learning to match features in learned invariant spaces \cite{sermanet2017time}. Finally, we note that much recent work has observed that the structure of observation sequences can be exploited to generate behavior, whether in the context of language modeling \cite{brown2020language}, 3D navigation \cite{dosovitskiy2017learning}, few-shot planning \cite{rybkin2019learning}, or value-based RL \cite{edwards2020estimating}. FORM uses generative models of future observations to exploit this property of observation transitions and connect it to inverse reinforcement learning to produce a practical algorithm for imitation.

\section{Approach}
\label{approach}

\subsection{Inverse reinforcement learning from observations}
Our goal is to learn a policy that produces behavior like an expert (or demonstrator) by IRL, using only observation. Historically, the IRL procedure has been framed as matching the expected distribution over states and actions (or their features) along the imitator and demonstrator paths \cite{ng2000algorithms, ng2004apprenticeship, ziebart2008maximum}. As also noted in \cite{arenz2020nonadversarial}, we can express this as a divergence minimization problem:
\begin{align}
\label{eq:irl_kl}
\min_\theta \mathcal{D}_{KL}[p_\theta^I(\tau) || p^D(\tau)],
\end{align}
where $\tau=\{x_0, a_0, x_1, a_1, \ldots, a_{T-2}, x_{T-1}\}$ is a trajectory consisting of actions are actions $A=\{a_0, \ldots a_{T-1}\}$ and states $X=\{x_0, \ldots, x_{T-1} \}$. We use $x$ rather than $o$ (for observation) or $s$ (for state) because FORM does not assume that its inputs are Markovian -- FORM applies to generic observations -- but speaking in terms of states simplifies the comparison to other methods (like GAIL) that assume Markovian states are given or inferred. $p^D(\tau)$ is the distribution over trajectories induced by the demonstrator's policy and the environment dynamics, while $p^I_{\theta}(\tau)$ is the corresponding distribution induced by an imitator with learnable parameters $\theta$.

In imitation learning from observation, the imitator must reason about the demonstrator's behavior without supervised access to the expert's actions (its control signals). Accordingly, we focus on distributions over observation sequences, which amounts to integrating out the imitator's actions:
\begin{align}
\nonumber
p_\theta^I(X) & =  \int_{A} p_\theta^I(\tau) =  \int_{A} p_\theta^I(A, X) \\
& = \int_{A} \prod_{t \ge 0} p(x_{t} | x_{<t}, a_{< t}) \pi_\theta^I(a_{t-1} | x_{<t}, a_{<t-1}).
\end{align}
This density reflects both the environment transition dynamics $p(x_{t} | x_{t-1}, a_{t-1})$ and the imitator policy $\pi_{\theta}(a_t | x_t)$, whose parameters $\theta$ we seek to learn. Similarly, we can write the probability of a demonstrator trajectory in terms of the unobserved expert policy as
\begin{align}
\nonumber
p^D(X) & = \prod_{t \ge 0} p^D(x_t | x_{<t}) \\
& = \int_{A} \prod_{t \ge 0} p(x_{t} | x_{<t}, a_{<t}) \pi^D(a_{t-1} | x_{<t}, a_{<t-1}).
\end{align}
Our objective is to minimize the KL-divergence\footnote{We use the reverse KL because the policy learns on its own trajectories, as in RL \cite{levine2018reinforcement}.} between these two densities:
\begin{align}
\label{eq:form_kl}
\nonumber
& \min_\theta \mathcal{D}_{KL}[p_\theta^I(X) || p^D(X)]
\\
& = \min_\theta \mathbb{E}_{p_\theta^I(X)} \bigg [ \log p_\theta^I(X)- \log p^D(X) \bigg].
\end{align}

Minimizing the divergence corresponds to maximizing the following expression in expectation:
\begin{align}
\label{eq:rho_form}
\mathcal{\rho}_\textsc{form} & = \log p^D(X) - \log p_\theta^I(X).
\end{align}
In this work, we propose to imitate by treating $\rho_\textsc{form}$ as a return and maximizing it using RL.

\subsection{Optimizing FORM with effect models}
\label{sec:optimizing_form}

To see how we will capture this expression, first note that each term of $\rho_\textsc{form}$ is a log-density over the states encountered in an episode $\log p(X)$, which we can rewrite as $\log p(x_0) + \sum_{t > 0} \log p(x_{t} | x_{<t})$ using the chain rule for probability. As the initial state is independent of the policy, we can simplify the expression used in each reward term to $\sum_{t \ge 0} \log p(x_t | x_{t-1})$. This means the return can be expressed solely in terms of next-step conditional densities. To simplify the discussion, we present all results from here forward in terms of one-step predictive models $\log p(x_{t} | x_{t-1})$, but the FORM derivation and algorithm applies equally well to generic sequence models $\log p(x_{t} | x_{<t})$.

We propose to learn a reward model by introducing models of the state transition densities under (1) the demonstrator $p^D(x_t | x_{t-1})$ and (2) the imitator $p^I_{\theta}(x_t | x_{t-1})$. We refer to these as \textit{effect models} to differentiate them from how ``model'' is used elsewhere in the RL literature to refer to models of transition dynamics (Figure \ref{fig:effect_model}). Unlike transition models, which are typically action-conditional and are assumed to model policy-independent transition dynamics, effect models are not conditioned on actions and attempt to capture the effects of policy and environment dynamics. A similar class of models was used to model an expert's behavior in recent work \cite{rhinehart2020deep}.

\begin{algorithm}[H]
\label{alg:form_mpo}
\SetAlgoLined
\kwInput{A fixed dataset $\mathbb{D}$ of expert state transitions, a replay buffer to fill with imitator data, an environment.}
\kwInit{Randomly initialize demonstrator effect model $p_\omega^D(x_t \mid x_{t-1})$, imitator effect model $p_\phi^I(x_t \mid x_{t-1})$, and imitator policy $\pi_\theta^I(a_t | x_t).$}
 \While{$p_\omega^D$ not converged}{
    \textit{\# Train demonstrator effect model}
    
    Sample batch of trajectories from the expert dataset $\mathbb{D}$. 
    
    Update $p_\omega^D$ by taking a gradient step (e.g. with Adam) on:
    \begin{align*}
        \max_\omega \mathbb{E}_{D} \bigg[ \sum_{t \ge 0} \log p_\omega^D(x_t \mid x_{t-1}) \bigg].
    \end{align*}
 }
 \While{$\pi_\theta^I(a_t | x_t)$ not converged}{
    \textit{\# Train imitator effect model and policy}
 
    Sample trajectories from the environment using $\pi_\theta^I$ and add them to the replay buffer. 
    
    Sample batch of trajectories from the replay buffer. Label the reward of each sampled transition $(x_{t-1}, a_{t-1}, x_t)$ using $p_\omega^D(x_t \mid x_{t-1})$ and $p_\phi^I(x_t \mid x_{t-1})$:
    \begin{align*}
        r_t = \log p_\omega^D(x_{t} \mid x_{t-1}) - \log p_\phi^I(x_t \mid x_{t-1}).
    \end{align*}    

    Update $\pi_\theta^I(a_t | x_t)$ with a step of a policy improvement algorithm (e.g. with MPO) using returns computed from the reward-labeled trajectories (e.g. with Retrace).

    Update $p_\phi^I(x_t \mid x_{t-1})$ by taking a gradient step (e.g. with Adam) on:
    \begin{align*}
        \max_\phi \mathbb{E}_{I}[\sum_{t \ge 0} \log p_\phi^I(x_t \mid x_{t-1})].
    \end{align*}
    
 }
 \caption{Imitation learning with FORM}
\end{algorithm}

We wish to maximize this return using standard tools for policy optimization. We can do this without introducing bias only if the policy gradients do not depend on gradients of any term in the reward (which aren't accounted for by standard policy optimizers). As we derive in Sec.~\ref{sec:policy_gradient} of the appendix, this assumption holds, and we can write the policy gradient as: 
\begin{align}
\label{eq:policy_grad_short}
\hspace{-3mm}\nabla_\theta \mathcal{J}_\textsc{form}(\pi^I_{\theta}) = \mathbb{E}_{\tau \sim \pi^I_{\theta}} [\rho_\textsc{form} \sum_{t \ge 0} \nabla_\theta \log \pi^I_\theta (a_t | x_t)].
\end{align}
Intuitively, the policy gradient does not involve gradients of either $p^I_\theta$ or $p^D$ because neither of these densities are conditioned on the actions sampled from the policy (in effect, the contribution of the density to the policy gradient is integrated out).
Because the demonstrator effect model is independent of the imitator, we can train it offline on expert demonstrations using a maximum likelihood objective:
\begin{align}
\max_\omega \mathbb{E}_{p^D(X)}[\sum_{t \ge 0} \log p_\omega^D(x_t \mid x_{t-1})].
\end{align}
The model of the imitator density $\log p_\theta^I(X)$, on the other hand, needs to capture the transition density under the current policy (it acts as a self-model). Accordingly, we train it by taking stochastic gradient descent steps on the following objective at the same time as the imitator policy is training:
\begin{align}
\max_\phi \mathbb{E}_{p_\theta^I(X)}[\sum_{t \ge 0} \log p_\phi^I(x_t \mid x_{t-1})].
\end{align}
By incorporating both models, we obtain the full FORM policy objective:
\begin{align}
\max_\theta \mathbb{E}_{\pi_\theta^I(X)}\bigg[\sum_{t \ge 0} \log p_\omega^D(x_{t} \mid x_{t-1}) - \log p_\phi^I(x_t \mid x_{t-1})\bigg].
\vspace{-3mm}
\end{align}

Despite the inclusion of two terms with opposite signs, the FORM policy objective is \textbf{not} an adversarial loss: FORM is based on a KL-minimization objective, rather than an adversarial minimax objective, and is not formulated as a zero-sum game. The second term in the objective can be viewed as an entropy-like expression, similar to the one that arises in maximum-entropy RL \cite{levine2018reinforcement}.

This objective includes both an expectation with respect to the current imitator policy and a term that reflects the current imitator effect model. This suggests that this objective is easiest to optimize in an on-policy setting. Nonetheless, we find that it can be stably optimized in a moderately off-policy setting. In all experiments here, we sample transitions from a replay buffer, computing rewards as they are consumed. We compute returns using the Retrace algorithm on the raw rewards \cite{munos2016safe} (which corrects for mildly off-policy actions using importance sampling). We optimize the policy using the MPO algorithm \cite{abdolmaleki2018maximum}. We choose MPO because it is known to perform well in mildly off-policy settings: FORM itself does not make any MPO-specific assumptions, and we expect it to perform well with many other policy optimizers. We describe our full procedure in Algorithm 1. 

\definecolor{Gray}{gray}{0.9}
\newcommand{\err}[1]{{\tiny $\pm$ #1}}
\begin{table*}[t]
\small
\begin{center}
\begin{tabular}{|l|l|l||l|l|l|l|l||l|}
\hline
               & Expert & BC               & BCO                       & GAIfO                      & GAIfO+GP                   & VAIfO                      & VAIfO+GP                   & FORM                      \\ \hline \hline
Reacher Easy   & 974.6  & 970.3 \err{12.2} & \textbf{966.6 \err{7.9}}  & 869.9 \err{48.6}           & 915.9 \err{37.8}           & 861.6 \err{61.4}           & 901.3 \err{30.4}           & \underline{950.2 \err{14.9}}          \\ \hline 
\rowcolor{Gray}
Reacher Hard   & 981.3  & 892.4 \err{19.1} & \underline{940.1 \err{3.9}}           & 818.7 \err{11.3}           & 783.7 \err{119.7}          & 604.4 \err{426.3}          & 891.0 \err{73.9}          & \textbf{957.3 \err{6.1}}  \\ \hline
Cheetah Run    & 930.5  & 227.5 \err{37.4} & 75.7  \err{4.2}           & 607.6 \err{429.6}          & \textbf{921.3 \err{6.9}}   & 820.0 \err{98.8}           & \underline{918.3 \err{6.4}}   & 827.9 \err{31.9}          \\ \hline
\rowcolor{Gray}
Quadruped Walk & 972.4  & 752.1 \err{37.3} & 191.9 \err{33.6}          & 672.6 \err{409.8}          & \underline{963.6 \err{4.8}}            & 927.8 \err{5.0}            & 945.8 \err{15.5}            & \textbf{963.6 \err{2.5}}  \\ \hline
Quadruped Run  & 962.9  & 719.2 \err{14.0} & 271.4 \err{48.4}          & 952.5 \err{7.5}            & \textbf{952.3 \err{2.1}}   & 926.6 \err{38.3}           & \underline{950.0 \err{2.7}}   & 948.5 \err{1.5}           \\ \hline
\rowcolor{Gray}
Hopper Stand   & 965.8  & 534.1 \err{13.2} & 91.4  \err{8.5}           & 400.0 \err{164.3}          & 748.5 \err{224.1}          & \underline{835.8 \err{103.0}}          & \textbf{891.2 \err{42.1}}          & 815.7 \err{9.2}  \\ \hline
Hopper Hop     & 711.5  & 98.4  \err{4.8}  & 9.1   \err{7.2}           & 689.2 \err{10.0}           & \textbf{694.4 \err{0.3}}   & \underline{610.5 \err{74.8}}           & 683.6 \err{22.3}   & 636.2 \err{38.9}          \\ \hline
\rowcolor{Gray}
Walker Stand   & 993.6  & 731.7 \err{29.7} & 385.9 \err{27.6}          & \underline{989.4 \err{1.5}}   & 985.4 \err{1.6}            & \textbf{989.4 \err{0.5}}   & 986.0 \err{1.9}            & 985.1 \err{2.6}           \\ \hline
Walker Walk    & 983.2  & 719.5 \err{50.0} & 61.9  \err{20.7}          & 976.5 \err{2.8}            & \textbf{981.6 \err{1.4}}   & 971.2 \err{5.6}            & 975.2 \err{1.3}   & \underline{977.8 \err{1.0}}           \\ \hline
\rowcolor{Gray}
Walker Run     & 952.1  & 108.5 \err{33.2} & 39.0  \err{7.8}           & \textbf{949.5 \err{5.6}}   & 947.6 \err{5.5}            & \underline{949.0 \err{2.6}}   & 948.5 \err{2.1}            & 942.0 \err{4.5}           \\ \hline
Humanoid Stand & 905.9  & 780.7 \err{30.5} & 9.99  \err{2.51}          & 4.9   \err{1.0}            & \underline{856.2 \err{15.5}}  & 257.5   \err{12.4}         & \textbf{863.5 \err{7.7}}  & 704.6 \err{12.1}          \\ \hline
\rowcolor{Gray}
Humanoid Walk  & 809.5  & 293.9 \err{16.2} & 9.61  \err{5.73}          & 1.2   \err{0.4}            & \textbf{798.4 \err{1.0}}   & 658.2   \err{123.6}        & \underline{795.5 \err{3.4}}   & 783.0 \err{3.3}           \\ \hline
Humanoid Run   & 736.6  & 54.2  \err{5.1}  & 1.04  \err{0.24}          & 0.6   \err{0.0}            & 683.4 \err{6.9}            & 676.6   \err{25.8}         & \textbf{691.6 \err{24.0}}            & \underline{691.1 \err{7.8}}  \\ \hline
\end{tabular}
\caption{Asymptotic performance on 13 tasks from six DCS domains (mean $\pm$ standard deviation across three seeds) of our method (FORM) and baselines Behavioral Cloning from Observations (BCO) \cite{torabi2018behavioral}, GAIL from Observations (GAIfO) \cite{torabi2019generative}, and regularized variants with a tuned gradient penalty \cite{gulrajani2017improved} (GAIfO+GP), a variational discriminator bottleneck \cite{peng2019variational} (VAIfO), or both forms of regularization (VAIfO+GP). Because BC \cite{pomerleau1989alvinn} uses expert actions it is not comparable to the other methods, but nevertheless performs poorly on many tasks, even with 1000 demonstrations. FORM performs competitively with well-regularized forms of GAIfO, while generally outperforming BCO and GAIfO. For each task, we highlight the method with \textbf{best} and \underline{second best} mean performance.}
\vspace{-6mm}
\label{table:asymptotic_performance}
\end{center}
\end{table*}

\vspace{-2mm}
\subsection{GAIL, occupancy-based imitation, and robustness}
\label{sec:gail_occupancy}
GAIL and its variants are justified in terms of matching the state-action occupancy of an expert -- GAIL attempts to unconditionally match the rates at which states and actions are visited -- rather than directly matching a policy or its effects. In contrast, FORM's reward is derived directly from an objective that matches a policy's effects on a sequence (eq. \ref{eq:rho_form}). This has consequences for their robustness, as we will explain. 

First, note that a policy is a local concept (it describes how to map states or observations to actions), while an occupancy is a global concept (it describes the rates at which an agent visits states and actions in expectation). To see why the occupancy is global, note that the occupancy \cite{ho2016generative, torabi2019generative} of a state $x_i$ by a policy $\pi$ is given by $\rho_{\pi}(x_i) = \sum_{t=0}^{\infty} \gamma^{t} p(x_t=x_i | \pi),$ where in general:

\begin{align}
p(x_t=x_i | \pi) = \int\displaylimits_{\{x, a\}_{<t}} p(x_t=x_i | x_{<t}, a_{<t}) p(x_{<t}, a_{<t} | \pi)
\end{align}

In other words, to reason about the occupancy of a state is to reason about every possible way the policy might arrive there. In practice, for GAIL, the discriminator computes a state's reward by comparing the frequency of $x_t$ to the frequency of all other states that are seen in the data, whatever the conditions under which that state was produced. Because FORM relies on conditional probabilities and does not depend on long-horizon visitation in its derivation, the only relevant states are those that appear under similar conditions. Essentially, FORM's reward involves comparisons to fewer observations because it takes a state's context -- namely, the transition that produced it -- into account. 

We expect this property to mitigate GAIL's sensitivity to noise. It's easiest to see why this should happen by comparing GAIfO and FORM for two-state inputs. Here, FORM maximizes $\log p^D(x_t | x_{t-1}) - \log p^I(x_t | x_{t-1})$ (each term estimated separately by maximum likelihood), while GAIfO maximizes $\log \frac{p^D(x_t, x_{t-1})}{p^I(x_t, x_{t-1})} = \log \frac{p^D(x_t | x_{t-1}) p^D(x_{t-1})}{p^I(x_t | x_{t-1}) p^I(x_{t-1})}$ (the entire log ratio is estimated in one go by a discriminator). If a feature is present in the imitator data but was never in the demonstrator data, then $p^D(x_{t-1})$ will be close to 0 on this data, driving the log ratio to $-\infty$ regardless of the probability of the transition that follows. The presence of noise makes spurious features like this inevitable. This makes it difficult for GAIL to focus on the meaningful controllable differences in the data, namely in the transition probabilities $p(x_t | x_{t-1})$. By estimating each term separately (avoiding a discriminator) and including only transition-related terms (using a conditional density), FORM reduces the susceptibility to sensitivity of this kind.

Finally, we note that GAIL is typically justified by the observation that recovering an expert's occupancy is equivalent to recovering its policy, but this is only true in Markov Decision Processes (MDPs) \cite{syed2008apprenticeship, ho2016generative} and not in general. In practice, imitation must often be done using noisy or high-dimensional observations rather than ground-truth MDP states, and matching occupancy in these spaces is problematic. In settings like this, relying on the global occupancy induced by a policy rather than on the immediate effects of a policy may lead to misleading results. For example, GAIL will attempt to match the occupancy of all noise dimensions, and this is usually possible. In practice, this means that the GAIL objective needs to be carefully regularized to avoid overfitting to irrelevant differences. These effects appear to be stronger when training an IRL agent from replay, as discussed in \cite{kostrikov2019discriminator}, and they may be further exacerbated when imitating without actions. In our experiments, GAIfO fails completely on the Humanoid tasks of the Control Suite when unregularized. Even with strong regularization, GAIfO is very sensitive to the presence of irrelevant differences between demonstrator and imitator domains, as our experiments illustrate.

\vspace{-3mm}
\section{Experiments}
\label{experiments}

We evaluate FORM against strong baselines on 13 tasks from six domains from the DeepMind Control Suite (DCS) \cite{tassa2018deepmind}, a set of benchmarks for continuous control domains, chosen to match those frequently used in the imitation learning literature\footnote{We note that many imitation learning methods are evaluated on the superficially similar OpenAI Gym Mujoco benchmark \cite{brockman2016gym}, but the Gym domains have essentially deterministic initial states and other properties that make them poorly suited for evaluating imitation learning methods (see Sec.~\ref{sec:gym_vs_dmc} of the Appendix for a discussion).}. All approaches use internal Mujoco state representations: these are smaller than e.g. image observations, and vary in size from 6- (reacher) to 67- (humanoid) and 78-dimensional (quadruped). As observed by \cite{zolna2020taskrelevant}, GAIL struggles to imitate in the presence of a small number of differences between expert and imitator domains. We conduct a similar experiment to characterize the robustness of FORM and GAIfO to irrelevant, but undersampled, factors of variation in the demonstrator data. Because the focus of our evaluation concerns robustness to distractors, rather than the minimum number of demonstrations needed for successful imitation, we conduct all experiments using 1000 demonstrations, sufficient to ensure mostly satisfactory performance in the absence of distractors.

\noindent \textbf{Expert data.} For all domains, we train an expert via RL on the ground truth task reward. Experts are trained to convergence using MPO, with the same policy and value architecture used for imitation (under all imitation conditions). For imitation, we generate a fixed dataset of 1000 demonstration trajectories from each policy, each of which depicts a single episode 1000 timesteps in duration (i.e.~$10^6$ steps total). All imitation methods are trained using the same demonstrator data.

\begin{figure*}[t]
    \centering
    \includegraphics[keepaspectratio,width=1.0\linewidth]{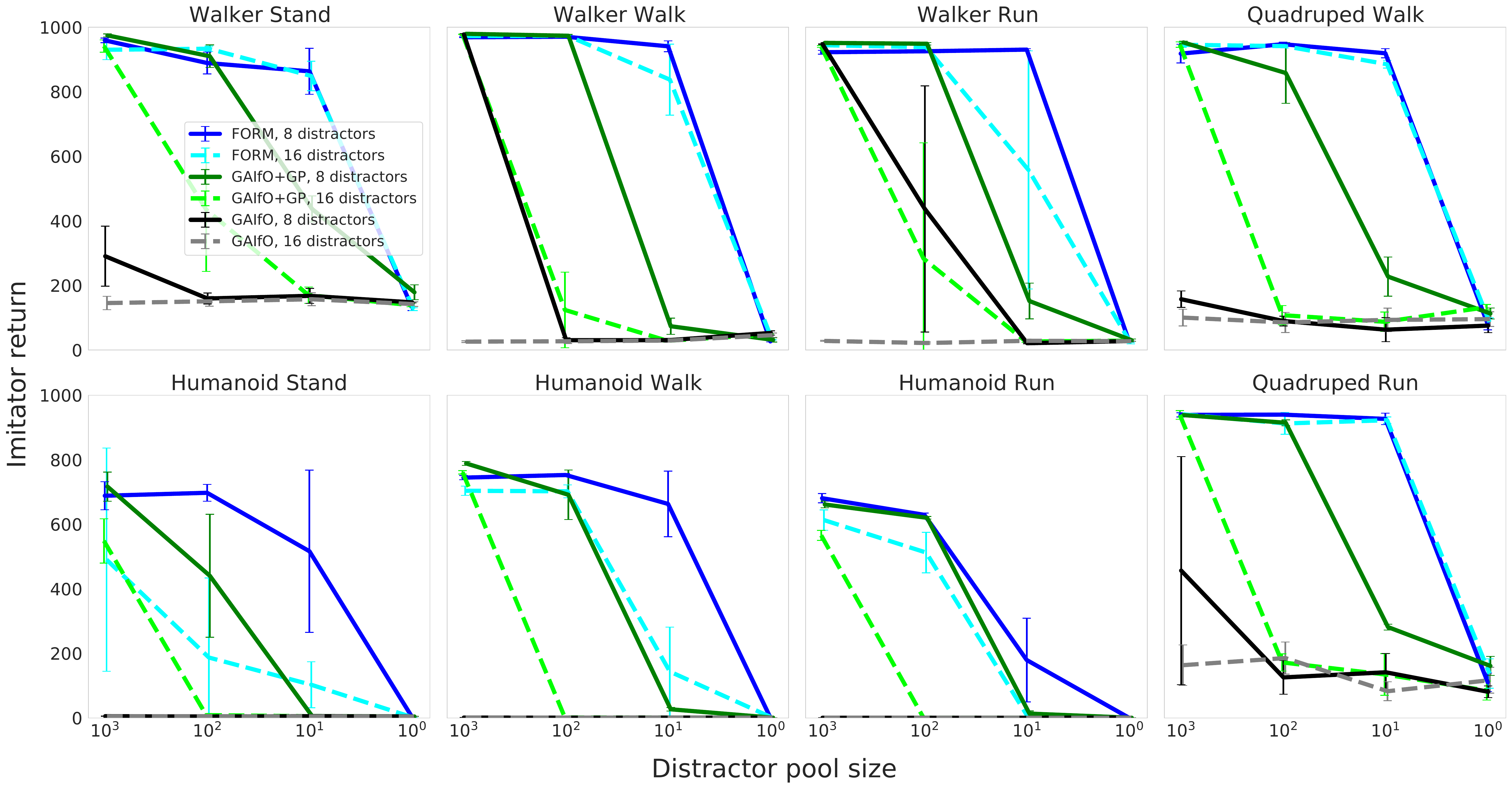}
    \vspace{-8pt}
    \caption{Performance of FORM, GAIfO+GP, and GAIfO in the presence of distractor features. The distractor pool size $M$ (\#unique points sampled in expert data) is varied from $10^3$ down to $1$ for both $N=8$ and $N=16$ distractor dimensions. FORM exhibits greater stability than GAIfO+GP in both settings, maintaining performance down to $M=10$. Error bars indicate standard deviation across 3 seeds. For reasons of legibility and space, comparison to all other baselines are given in the appendix.}
    \label{fig:distractors}
\end{figure*}

\noindent \textbf{Distractor data.} To probe robustness to a domain shift between the expert and imitation domains, we deliberately introduce spurious signals, unrelated to the task or agent state, into the state observation vector. During the demonstration phase, these take the form of binary noise patterns drawn from a fixed set which are appended to the state vector and held constant for the duration of the episode. During imitation binary noise vectors are also appended, but any binary pattern is permitted (i.e.~no longer need come from the fixed set). The different patterns appended during the expert and imitation phases impose a domain shift between them.

Formally, during imitation we append the state vector $x_t$ with a binary pattern $b \sim [0,1]^N$ to form an augmented observation $\tilde{x}_t = [x_t;b]$, where $N$ is the number of distractor dimensions. During demonstration, $b \sim \{b_1,\ldots,b_M\}, b \in [0,1]^N$, while $M$ controls the the number of distinct patterns, known as the \emph{pool size}. $M$ and $N$ control the magnitude of the domain shift: increasing $N$ makes the task harder by reducing the fraction of state that contains signal, while increasing $M$ makes the task easier by ensuring that all distractor features are present in both demonstrator and imitator data.

Due to the input normalization procedure (Sec.~\ref{sec:form_details}), the IL agent has no way of distinguishing noise dimensions from ones carrying state information. Ideally however, it should learn to ignore the extra dimensions since they are unrelated to the task, making it robust to changes in the distractor pattern. Our setup directly parallels situations encountered in practice involving undersampled factors of variation. For example, when performing IL using visual inputs with a robot, the background appearance of the rooms in which the expert data collection and imitation during deployment are performed correspond to two distinct distractor patterns that are intermingled with task-relevant portions of the state. For IL to work in such settings the algorithm must be robust to changes in the background distractors. We can see how sensitive a model is to the presence of undersampled factors of variation by observing how stable its performance is as the pool size $M$ decreases.

\subsection{Details of the FORM implementation}
\label{sec:form_details}
\textbf{Architecture.} We use simple feedforward architectures to parameterize the density models (3 layer MLPs with 256 units, and tanh and ELU \cite{djorkarne2016fast} nonlinearities). We model the density as a mixture of 4 Gaussian components, with the network outputting GMM mixture coefficients and the means and standard deviations of each component. We use Gaussians with a diagonal covariance matrix. In all experiments, we clip the standard deviation to a minimum value of  0.0001. We use the same architecture and same hyperparameters for the imitator and demonstrator effect model in each setting.

\textbf{Effect model training.} All demonstrator models were trained offline for 2 million steps. We standardized effect model inputs using per-dimension means and variances estimated by by exponential moving average: we found that this improved generative model training (it did not affect GAIfO training). 

Three forms of regularization were used with the demonstrator and imitator generative models:
(i) $\ell_2$ weight-decay, (ii) training on data generated by agent rollouts, i.e. using the network output at a timestep as the input at the next during training (a common trick used in the recurrent neural network literature \cite{bengio2015scheduled}), (iii) prediction of observations at multiple future timesteps \cite{hafner2019learning}. See the Appendix for more details. 
In all experiments, we share the hyperparameter settings of all regularizers between the demonstrator and imitator effect model (we do not tune them separately). We tuned $\ell_2$ weight (sweeping values of $[0.0, 0.01, 0.1,$ and $1.0]$) and the fraction of each batch generated by agent rollouts (sweeping values of $[0.0, 0.01, 0.1, 1.0]$) per domain, but otherwise use identical hyperparameters for all FORM models.

\subsection{Baselines: GAIfO and BCO}

To ensure DMCS experiments were fair and well-calibrated, we impelemented and tuned a strong GAIfO baseline. The GAIfO discriminator is conditioned on the current observation. We found that there was no benefit to conditioning on pairs of subsequent observations (see Table~\ref{table:gail_frames} in the Appendix). This is likely because DCS observations include velocity observations as well as static positions. The discriminator network uses the same architecture as the FORM effect models except for the mixture-of-Gaussians head, which is replaced by a (scalar) classifier head. We additionally found that there was no benefit in standardizing the observations as we do for FORM. For GAIfO+GP, we apply a gradient penalty \cite{gulrajani2017improved} to the last two layers of the discriminator. For VAIfO (VAIL from observations), we introduce a variational bottleneck in the discriminator architecture and add a KL-constraint term to the loss, as in \cite{peng2019variational}. Following \cite{kostrikov2019discriminator}, we train both the policy and the discriminator using data sampled from a replay buffer.

In BCO, an inverse model $p(a_t | x_t, x_{t+1})$ is trained on imitator trajectories and then used to label the actions on demonstrator trajectories \cite{torabi2018behavioral}. We train the inverse model using the same architecture as the FORM effector models and the GAIfO discriminator, replacing the output head with a Gaussian distribution (the same class of distributions used by the RL agent to produce the actions). The BCO agent is then trained in a supervised fashion on expert trajectories labeled by the inverse model. The BC agent is trained directly on expert trajectories with expert actions. Because BC is trained using expert actions, while the other imitation algorithms we evaluate are not, it is not strictly comparable. We include it to calibrate readers to the difficulty of these tasks and the relative performance of the algorithms we evaluate for imitation from observation.

\begin{figure*}[h!]

    \centering
    
    \begin{minipage}[b]{0.32\linewidth}
    \includegraphics[width=\linewidth]{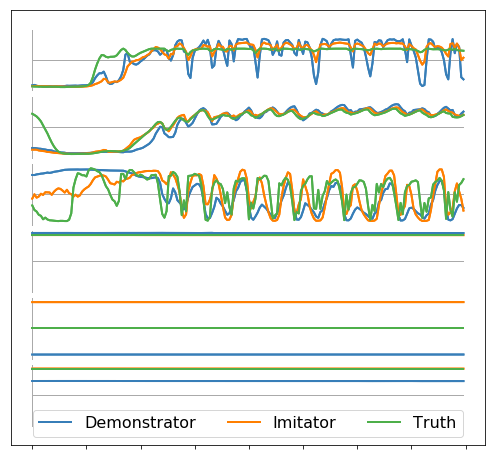} \\ \vspace{-6.5mm}\center{Distractor 1}
    \end{minipage}
    \begin{minipage}[b]{0.32\linewidth}
    \includegraphics[width=\linewidth]{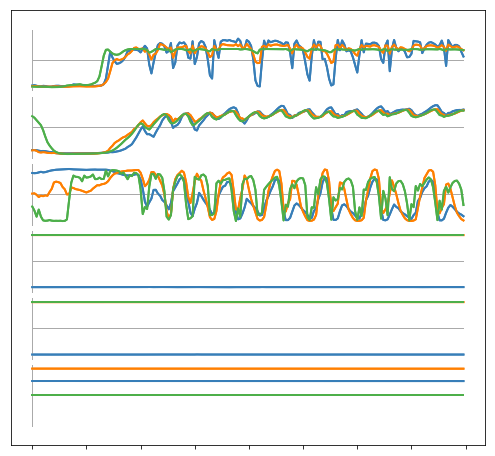}\\ \vspace{-6.5mm}\center{Distractor 2}
    \end{minipage}
    \begin{minipage}[b]{0.32\linewidth}
        \includegraphics[width=\linewidth]{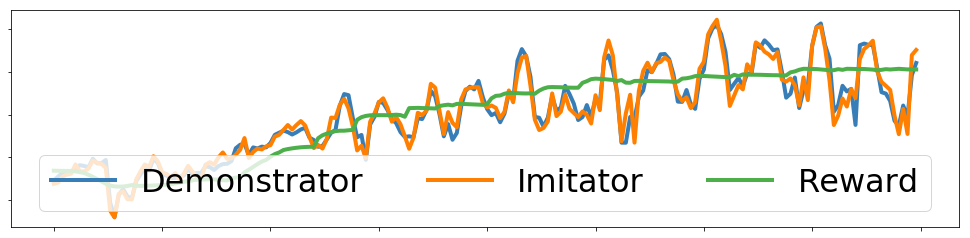}
        \includegraphics[width=\linewidth]{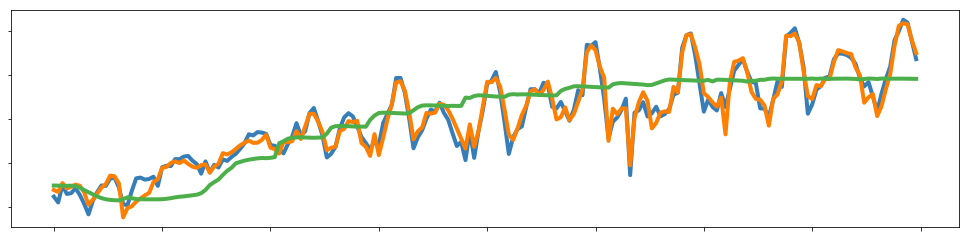}
        \includegraphics[width=\linewidth]{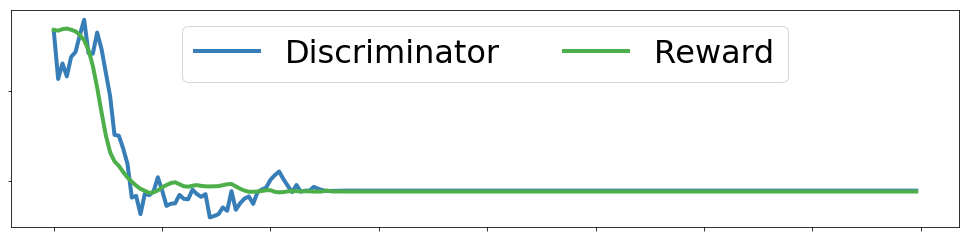}
        \includegraphics[width=\linewidth]{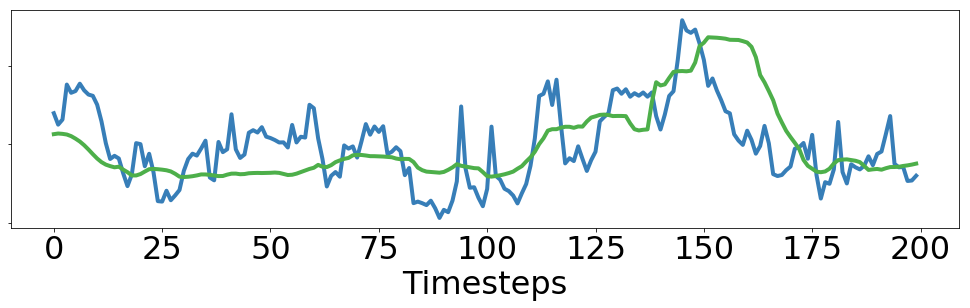}
    \end{minipage}

    \caption{
    \textbf{Left, Middle} --- subset of observation dimensions (3 internal states (top) and 3 distractor features (bottom)) from a FORM imitation agent on the Walker Run task. Distractor features appear as horizontal lines as they do not vary with time in an episode. \textcolor{darkgreen}{True observations}, with \textcolor{blue}{demonstrator} and \textcolor{orange}{imitator} predictions overlaid. 
    Distractor 1 shows the agent learning with a pattern previously seen in the expert demonstrations. Distractor 2 uses a novel pattern not seen in expert data. 
    The FORM agent behavior and model predictions are qualitatively unchanged, showing robustness to the distractor pattern.
    \textbf{Right} --- reward model traces for both FORM and GAIfO+GP,
    alongside the ground truth \textcolor{darkgreen}{task reward}.
    \emph{Top}: the log-likelihoods of the \textcolor{blue}{demonstrator} and \textcolor{orange}{imitator} components of the FORM reward model, for distractor 1 (top) and distractor 2 (bottom).
    \emph{Bottom}: the expert probability output by the GAIfO+GP \textcolor{blue}{discriminator} for distractor 1 (top) and distractor 2 (bottom).
    Both FORM and GAIfO+GP agents were trained with with $N=8$ distractor dimensions and a pool of $M=10$ distractors in expert data. 
    FORM is robust to distractor features in this setting even when its predictions are imperfect, and the imitation agent obtains good reward on the task. In contrast, the behavior of the GAIfO+GP agent depends significantly on the distractors and largely fails at the task. 
    }
    \label{fig:traces}
\end{figure*}

\subsection{Policy architecture}

For both IRL methods (FORM and GAIfO), the underlying policy is trained with MPO and experience replay. Both the policy and critic networks encode a concatenation of the environment's observations that has been 
passed through a tanh activation. Both encode the observations with independent 3-layer MLPs using ELU activations. The policy network then projects to parameterize the mean and scale of a Gaussian action distribution. The critic concatenates the sampled action, applies layernorm \cite{ba2016layer} and a tanh, and applies another 3-layer MLP to produce the $Q$-value. All hidden layers have a width of 256 units.

\vspace{-2mm}
\subsection{Results}
\label{sec:results}
\textbf{No distractors.} In Table \ref{table:asymptotic_performance} we compare FORM to BCO and GAIfO, with various strong forms of regularization on the DeepMind Control Suite in the absence of distractors. The results are shown alongside the reward obtained by the expert RL agent. BCO succeeds only on the Reacher domain, performing poorly on the others. GAIfO in general performs well but fails completely on the Humanoid domain. The addition of a tuned gradient penalty or the introduction of a variational discriminator bottleneck allows GAIfO to also perform well on the Humanoid domain. FORM achieves competitive performance to strongly regularized forms of GAIfO (GAIfO+GP, VAIfO, VAIfO+GP). Despite access to 1000 demonstrations, no method is able to match expert performance on the Humanoid tasks, illustrating the challenge this domain poses due to the dimensionality of the state-space and highly variable initial conditions. This number of demonstrations (1000) may seem large when compared to the numbers used in work that uses the Gym benchmark \cite{brockman2016gym}. But please note that tasks from the Gym benchmark are easier to imitate, requiring almost no generalization between demonstrator and imitator due to its essentially deterministic initialization.

\textbf{With distractors.} We now explore the robustness of the different approaches to settings where distractors are present in the observations. Figure \ref{fig:distractors} compares FORM with GAIfO and GAIfO+GP with $N=8$ and $N=16$ dimensions of distractor features, as the distractor pool size $M$ is varied from $10^3$ down to $1$.  With $N=8$ distractor dimensions, FORM is consistently able to maintain performance as $M=10$, by which point the performance of GAIfO+GP has dropped significantly. For $N=16$ distractor dimensions the degradation for GAIfO+GP is more severe, even at the easier setting of $M=100$. In contrast, FORM is still able to perform well on most tasks (Humanoid Stand being the exception). We compare FORM to all other baselines on this setting in Figs.~\ref{fig:distractors_bc} and~\ref{fig:distractors_vail} of the appendix, and this trend holds generally. We find that GAIfO with a variational bottleneck regularizer (VAIfO and VAIfO+GP) performs similarly to GAIfO, and still exhibits sensitivity to noise. BC and BCO generally perform worse than FORM, but BC shows good noise resilience on Humanoid Stand in particular.

Figure \ref{fig:traces} visualizes the effect on imitation performance when the distractor pattern ($M=10$, $N=8)$ is changed between imitation training runs on the Walker Run task. Demonstrator and imitator model predictions from the FORM model show minimal change, with the agent achieving good reward for both patterns. In contrast, the GAIfO+GP model is highly sensitive to the change in distractor pattern and fails at the task. Collectively, these results show the fragility of GAIL-based IL methods to task-irrelevant features, and also illustrate the superior robustness of FORM in this setting. 

\section{Discussion}
\vspace{-2mm}
\label{conclusion}
In this work we introduce the Future Observation Reward Model, or FORM, an approach to inverse reinforcement learning that can be used for imitation from observations without actions. FORM makes few assumptions about the data being modeled, which makes it a promising approach for learning behavior from data collected under realistic conditions. In particular, we show that FORM is competitive with GAIL from observations while exhibiting improved stability in the face of spurious features. FORM imitates using likelihood-based generative models, a family of models that has been extensively studied and that can be scaled to real-world, noisy data. These properties make FORM a good candidate for the development of sophisticated approaches to imitation that can handle high-dimensional data with domain shifts.

FORM currently has several limitations. The demonstrator model $p^D$ must be trained off-line before learning the imitator model $p^I$ and policy $\pi^I$. This two-stage training is inefficient in wall clock terms relative to the monolithic training procedure of GAIL. This is compounded by the difficulty in assessing the quality of $p^D$ using training likelihood alone. In practice, we find that it is a poor predictor of subsequent imitator performance, necessitating both both training stages to be performed in order to ascertain if $p^D$ was modeled effectively. A second issue is that we currently model proprioceptive state: moving to image pixel-based inputs will require larger and more complex generative models, which will likely lead to added difficulties. 

\section*{Acknowledgements}

We are grateful to Josh Abramson, Feryal Behbahani, Federico Carnevale, Ashley Edwards, Tom Erez, Karol Gregor, Raia Hadsell, Leonard Hasenclever, Nicolas Heess, Alden Hung, Josh Merel, Nikolay Savinov, Yuval Tassa, Konrad Zolna and others at DeepMind for insightful discussions and suggestions.

\bibliography{bibliography}

\begin{thebibliography}{66}
\providecommand{\natexlab}[1]{#1}
\providecommand{\url}[1]{\texttt{#1}}
\expandafter\ifx\csname urlstyle\endcsname\relax
  \providecommand{\doi}[1]{doi: #1}\else
  \providecommand{\doi}{doi: \begingroup \urlstyle{rm}\Url}\fi

\bibitem[Abbeel \& Ng(2004)Abbeel and Ng]{ng2004apprenticeship}
Abbeel, P. and Ng, A.~Y.
\newblock Apprenticeship learning via inverse reinforcement learning.
\newblock In \emph{Proceedings of International Conference on Machine Learning
  (ICML)}, 2004.

\bibitem[Abbeel et~al.(2010)Abbeel, Coates, and Ng]{abbeel2010autonomous}
Abbeel, P., Coates, A., and Ng, A.~Y.
\newblock Autonomous helicopter aerobatics through apprenticeship learning.
\newblock \emph{International Journal of Robotics Research}, 29\penalty0
  (13):\penalty0 1608–1639, 2010.

\bibitem[Abdolmaleki et~al.(2018)Abdolmaleki, Springenberg, Tassa, Munos,
  Heess, and Riedmiller]{abdolmaleki2018maximum}
Abdolmaleki, A., Springenberg, J.~T., Tassa, Y., Munos, R., Heess, N., and
  Riedmiller, M.
\newblock Maximum a posteriori policy optimisation.
\newblock In \emph{Proceedings of International Conference on Learning
  Representations (ICLR)}, 2018.

\bibitem[Abramson et~al.(2020)Abramson, Ahuja, Brussee, Carnevale, Cassin,
  Clark, Dudzik, Georgiev, Guy, Harley, Hill, Hung, Kenton, Landon, Lillicrap,
  Mathewson, Muldal, Santoro, Savinov, Varma, Wayne, Wong, Yan, and
  Zhu]{abramson2020imitating}
Abramson, J., Ahuja, A., Brussee, A., Carnevale, F., Cassin, M., Clark, S.,
  Dudzik, A., Georgiev, P., Guy, A., Harley, T., Hill, F., Hung, A., Kenton,
  Z., Landon, J., Lillicrap, T., Mathewson, K., Muldal, A., Santoro, A.,
  Savinov, N., Varma, V., Wayne, G., Wong, N., Yan, C., and Zhu, R.
\newblock Imitating interactive intelligence.
\newblock \emph{arXiv preprint arXiv:2012.05672}, 2020.

\bibitem[Arenz \& Neumann(2020)Arenz and Neumann]{arenz2020nonadversarial}
Arenz, O. and Neumann, G.
\newblock Non-adversarial imitation learning and its connections to adversarial
  methods.
\newblock \emph{arXiv preprint arXiv:2008.03525}, 2020.

\bibitem[Arjovsky et~al.(2017)Arjovsky, Chintala, and
  Bottou]{arjovsky2017wasserstein}
Arjovsky, M., Chintala, S., and Bottou, L.
\newblock Wasserstein {GAN}.
\newblock In \emph{Proceedings of International Conference on Machine Learning
  (ICML)}, 2017.

\bibitem[Ba et~al.(2016)Ba, Kiros, and Hinton]{ba2016layer}
Ba, J.~L., Kiros, J.~R., and Hinton, G.~E.
\newblock Layer normalization.
\newblock \emph{arXiv preprint arXiv:1607.06450}, 2016.

\bibitem[Babuschkin et~al.(2020)Babuschkin, Baumli, Bell, Bhupatiraju, Bruce,
  Buchlovsky, Budden, Cai, Clark, Danihelka, Fantacci, Godwin, Jones, Hennigan,
  Hessel, Kapturowski, Keck, Kemaev, King, Martens, Mikulik, Norman, Quan,
  Papamakarios, Ring, Ruiz, Sanchez, Schneider, Sezener, Spencer, Srinivasan,
  Stokowiec, and Viola]{deepmind2020jax}
Babuschkin, I., Baumli, K., Bell, A., Bhupatiraju, S., Bruce, J., Buchlovsky,
  P., Budden, D., Cai, T., Clark, A., Danihelka, I., Fantacci, C., Godwin, J.,
  Jones, C., Hennigan, T., Hessel, M., Kapturowski, S., Keck, T., Kemaev, I.,
  King, M., Martens, L., Mikulik, V., Norman, T., Quan, J., Papamakarios, G.,
  Ring, R., Ruiz, F., Sanchez, A., Schneider, R., Sezener, E., Spencer, S.,
  Srinivasan, S., Stokowiec, W., and Viola, F.
\newblock The {D}eep{M}ind {JAX} {E}cosystem, 2020.
\newblock URL \url{http://github.com/deepmind}.

\bibitem[Baram et~al.(2016)Baram, Anschel, and Mannor]{baram2016model}
Baram, N., Anschel, O., and Mannor, S.
\newblock Model-based adversarial imitation learning.
\newblock In \emph{Proceedings of Neural Information Processing Systems
  (NeurIPS)}, 2016.

\bibitem[Bengio et~al.(2015)Bengio, Vinyals, Jaitly, and
  Shazeer]{bengio2015scheduled}
Bengio, S., Vinyals, O., Jaitly, N., and Shazeer, N.~M.
\newblock Scheduled sampling for sequence prediction with recurrent neural
  networks.
\newblock In \emph{Proceedings of Neural Information Processing Systems
  (NeurIPS)}, 2015.

\bibitem[Bradbury et~al.(2018)Bradbury, Frostig, Hawkins, Johnson, Leary,
  Maclaurin, Necula, Paszke, Vander{P}las, Wanderman-{M}ilne, and
  Zhang]{jax2018github}
Bradbury, J., Frostig, R., Hawkins, P., Johnson, M.~J., Leary, C., Maclaurin,
  D., Necula, G., Paszke, A., Vander{P}las, J., Wanderman-{M}ilne, S., and
  Zhang, Q.
\newblock {JAX}: composable transformations of {P}ython+{N}um{P}y programs,
  2018.
\newblock URL \url{http://github.com/google/jax}.

\bibitem[Brockman et~al.(2016)Brockman, Cheung, Pettersson, Schneider,
  Schulman, Tang, and Zaremba]{brockman2016gym}
Brockman, G., Cheung, V., Pettersson, L., Schneider, J., Schulman, J., Tang,
  J., and Zaremba, W.
\newblock Open{AI} gym.
\newblock \emph{arXiv preprint arXiv:1606.01540}, 2016.

\bibitem[Brown et~al.(2020)Brown, Mann, Ryder, Subbiah, Kaplan, Dhariwal,
  Neelakantan, Shyam, Sastry, Askell, Agarwal, Herbert-Voss, Krueger, Henighan,
  Child, Ramesh, Ziegler, Wu, Winter, Hesse, Chen, Sigler, Litwin, Gray, Chess,
  Clark, Berner, McCandlish, Radford, Sutskever, and Amodei]{brown2020language}
Brown, T., Mann, B., Ryder, N., Subbiah, M., Kaplan, J.~D., Dhariwal, P.,
  Neelakantan, A., Shyam, P., Sastry, G., Askell, A., Agarwal, S.,
  Herbert-Voss, A., Krueger, G., Henighan, T., Child, R., Ramesh, A., Ziegler,
  D., Wu, J., Winter, C., Hesse, C., Chen, M., Sigler, E., Litwin, M., Gray,
  S., Chess, B., Clark, J., Berner, C., McCandlish, S., Radford, A., Sutskever,
  I., and Amodei, D.
\newblock Language models are few-shot learners.
\newblock In \emph{Proceedings of Neural Information Processing Systems
  (NeurIPS)}, 2020.

\bibitem[Byrne(2009)]{laland2009animal}
Byrne, R.~W.
\newblock Animal imitation.
\newblock \emph{Current Biology}, 19\penalty0 (3):\penalty0 111 -- 114, 2009.

\bibitem[Clevert et~al.(2016)Clevert, Unterthiner, and
  Hochreiter]{djorkarne2016fast}
Clevert, D.-A., Unterthiner, T., and Hochreiter, S.
\newblock Fast and accurate deep network learning by exponential linear units
  ({ELUs}).
\newblock In \emph{Proceedings of International Conference on Learning
  Representations (ICLR)}, 2016.

\bibitem[Dosovitskiy \& Koltun(2017)Dosovitskiy and
  Koltun]{dosovitskiy2017learning}
Dosovitskiy, A. and Koltun, V.
\newblock Learning to act by predicting the future.
\newblock In \emph{Proceedings of International Conference on Learning
  Representations (ICLR)}, 2017.

\bibitem[Edwards et~al.(2018)Edwards, Sahni, Schroecker, and
  Isbell]{edwards2019imitating}
Edwards, A.~D., Sahni, H., Schroecker, Y., and Isbell, C.~L.
\newblock Imitating latent policies from observation.
\newblock In \emph{Proceedings of International Conference on Machine Learning
  (ICML)}, 2018.

\bibitem[Edwards et~al.(2020)Edwards, Sahni, Liu, Hung, Jain, Wang, Ecoffet,
  Miconi, Isbell, and Yosinski]{edwards2020estimating}
Edwards, A.~D., Sahni, H., Liu, R., Hung, J., Jain, A., Wang, R., Ecoffet, A.,
  Miconi, T., Isbell, C., and Yosinski, J.
\newblock Estimating {Q(s,s')} with deep deterministic dynamics gradients.
\newblock In \emph{Proceedings of International Conference on Machine Learning
  (ICML)}, 2020.

\bibitem[Fu et~al.(2018)Fu, Luo, and Levine]{fu2018learning}
Fu, J., Luo, K., and Levine, S.
\newblock Learning robust rewards with adversarial inverse reinforcement
  learning.
\newblock In \emph{Proceedings of International Conference on Learning
  Representations (ICLR)}, 2018.

\bibitem[Ghasemipour et~al.(2019)Ghasemipour, Zemel, and
  Gu]{ghasemipour2019divergence}
Ghasemipour, S. K.~S., Zemel, R., and Gu, S.
\newblock A divergence minimization perspective on imitation learning methods.
\newblock In \emph{Conference on Robotic Learning (CoRL)}, 2019.

\bibitem[Gulcehre et~al.(2020)Gulcehre, Paine, Shahriari, Denil, Hoffman,
  Soyer, Tanburn, Kapturowski, Rabinowitz, Williams, Barth-Maron, Wang,
  de~Freitas, and Team]{gulcehre2020making}
Gulcehre, C., Paine, T.~L., Shahriari, B., Denil, M., Hoffman, M., Soyer, H.,
  Tanburn, R., Kapturowski, S., Rabinowitz, N., Williams, D., Barth-Maron, G.,
  Wang, Z., de~Freitas, N., and Team, W.
\newblock Making efficient use of demonstrations to solve hard exploration
  problems.
\newblock In \emph{Proceedings of International Conference on Learning
  Representations (ICLR)}, 2020.

\bibitem[Gulrajani et~al.(2017)Gulrajani, Ahmed, Arjovsky, Dumoulin, and
  Courville]{gulrajani2017improved}
Gulrajani, I., Ahmed, F., Arjovsky, M., Dumoulin, V., and Courville, A.~C.
\newblock Improved training of wasserstein {GANs}.
\newblock In \emph{Proceedings of Neural Information Processing Systems
  (NeurIPS)}, 2017.

\bibitem[Hafner et~al.(2019)Hafner, Lillicrap, Fischer, Villegas, Ha, Lee, and
  Davidson]{hafner2019learning}
Hafner, D., Lillicrap, T., Fischer, I., Villegas, R., Ha, D., Lee, H., and
  Davidson, J.
\newblock Learning latent dynamics for planning from pixels.
\newblock In \emph{Proceedings of International Conference on Machine Learning
  (ICML)}, 2019.

\bibitem[Ho \& Ermon(2016)Ho and Ermon]{ho2016generative}
Ho, J. and Ermon, S.
\newblock Generative adversarial imitation learning.
\newblock In \emph{Proceedings of Neural Information Processing Systems
  (NeurIPS)}, 2016.

\bibitem[Huber et~al.(2009)Huber, Range, Voelkl, Szucsich, Vir\'{a}nyi, and
  Miklosi]{huber2009theevolution}
Huber, L., Range, F., Voelkl, B., Szucsich, A., Vir\'{a}nyi, Z., and Miklosi,
  A.
\newblock The evolution of imitation: what do the capacities of non-human
  animals tell us about the mechanisms of imitation?
\newblock \emph{Philosophical Transactions of the Royal Society of London B},
  364\penalty0 (1528):\penalty0 2299 – 2309, 2009.

\bibitem[Jarrett et~al.(2020)Jarrett, Bica, and van~der
  Schaar]{jarrett2020strictly}
Jarrett, D., Bica, I., and van~der Schaar, M.
\newblock Strictly batch imitation learning by energy-based distribution
  matching.
\newblock In \emph{Proceedings of Neural Information Processing Systems
  (NeurIPS)}, 2020.

\bibitem[Kingma \& Ba(2014)Kingma and Ba]{kingma2014adam}
Kingma, D.~P. and Ba, J.
\newblock Adam: A method for stochastic optimization.
\newblock In \emph{Proceedings of International Conference on Learning
  Representations (ICLR)}, 2014.

\bibitem[Kostrikov et~al.(2019)Kostrikov, Agrawal, Dwibedi, Levine, and
  Tompson]{kostrikov2019discriminator}
Kostrikov, I., Agrawal, K.~K., Dwibedi, D., Levine, S., and Tompson, J.
\newblock Discriminator-actor-critic: Addressing sample inefficiency and reward
  bias in adversarial imitation learning.
\newblock In \emph{Proceedings of International Conference on Learning
  Representations (ICLR)}, 2019.

\bibitem[Kostrikov et~al.(2020)Kostrikov, Nachum, and
  Tompson]{kostrikov2020imitation}
Kostrikov, I., Nachum, O., and Tompson, J.
\newblock Imitation learning via off-policy distribution matching.
\newblock In \emph{Proceedings of International Conference on Learning
  Representations (ICLR)}, 2020.

\bibitem[Laland(2008)]{laland2008animal}
Laland, K.
\newblock Animal cultures.
\newblock \emph{Current Biology}, 18\penalty0 (9):\penalty0 366 -- 370, 2008.

\bibitem[Levine(2018)]{levine2018reinforcement}
Levine, S.
\newblock Reinforcement learning and control as probabilistic inference:
  Tutorial and review.
\newblock \emph{arXiv preprint arXiv:1805.00909}, 2018.

\bibitem[Liu et~al.(2021)Liu, He, Xu, and Zhang]{liu2021energy}
Liu, M., He, T., Xu, M., and Zhang, W.
\newblock Energy-based imitation learning.
\newblock In \emph{Proceedings of the International Conference on Autonomous
  Agents and Multiagent Systems (AAMAS)}, 2021.

\bibitem[Mania et~al.(2018)Mania, Guy, and Recht]{mania2018simple}
Mania, H., Guy, A., and Recht, B.
\newblock Simple random search of static linear policies is competitive for
  reinforcement learning.
\newblock In \emph{Proceedings of Neural Information Processing Systems
  (NeurIPS)}, 2018.

\bibitem[Merel et~al.(2017)Merel, Tassa, TB, Srinivasan, Lemmon, Wang, Wayne,
  and Heess]{merel2017learning}
Merel, J., Tassa, Y., TB, D., Srinivasan, S., Lemmon, J., Wang, Z., Wayne, G.,
  and Heess, N.
\newblock Learning human behaviors from motion capture by adversarial
  imitation.
\newblock \emph{arXiv preprint arXiv:1707.02201}, 2017.

\bibitem[Merel et~al.(2019)Merel, Hasenclever, Galashov, Ahuja, Pham, Wayne,
  Teh, and Heess]{merel2019neural}
Merel, J., Hasenclever, L., Galashov, A., Ahuja, A., Pham, V., Wayne, G., Teh,
  Y.~W., and Heess, N.
\newblock Neural probabilistic motor primitives for humanoid control.
\newblock In \emph{Proceedings of International Conference on Learning
  Representations (ICLR)}, 2019.

\bibitem[Mescheder et~al.(2017)Mescheder, Nowozin, and
  Geiger]{mescheder2017numerics}
Mescheder, L., Nowozin, S., and Geiger, A.
\newblock The numerics of {GANs}.
\newblock In \emph{Proceedings of Neural Information Processing Systems
  (NeurIPS)}, 2017.

\bibitem[Munos et~al.(2016)Munos, Stepleton, Harutyunyan, and
  Bellemare]{munos2016safe}
Munos, R., Stepleton, T., Harutyunyan, A., and Bellemare, M.~G.
\newblock Safe and efficient off-policy reinforcement learning.
\newblock In \emph{Proceedings of Neural Information Processing Systems
  (NeurIPS)}, 2016.

\bibitem[Ng \& Russell(2000)Ng and Russell]{ng2000algorithms}
Ng, A. and Russell, S.
\newblock Algorithms for inverse reinforcement learning.
\newblock In \emph{Proceedings of International Conference on Machine Learning
  (ICML)}, 2000.

\bibitem[OpenAI et~al.(2019)OpenAI, Berner, Brockman, Chan, Cheung, Dębiak,
  Dennison, Farhi, Fischer, Hashme, Hesse, Józefowicz, Gray, Olsson, Pachocki,
  Petrov, de~Oliveira~Pinto, Raiman, Salimans, Schlatter, Schneider, Sidor,
  Sutskever, Tang, Wolski, and Zhang]{openai2019dota}
OpenAI, Berner, C., Brockman, G., Chan, B., Cheung, V., Dębiak, P., Dennison,
  C., Farhi, D., Fischer, Q., Hashme, S., Hesse, C., Józefowicz, R., Gray, S.,
  Olsson, C., Pachocki, J., Petrov, M., de~Oliveira~Pinto, H.~P., Raiman, J.,
  Salimans, T., Schlatter, J., Schneider, J., Sidor, S., Sutskever, I., Tang,
  J., Wolski, F., and Zhang, S.
\newblock Dota 2 with large scale deep reinforcement learning.
\newblock \emph{arXiv preprint arXiv:1910.07113}, 2019.

\bibitem[Osa et~al.(2018)Osa, Pajarinen, Neumann, Bagnell, Abbeell, and
  Peters]{osa2018algorithmic}
Osa, T., Pajarinen, J., Neumann, G., Bagnell, J.~A., Abbeell, P., and Peters,
  J.
\newblock An algorithmic perspective on imitation learning.
\newblock \emph{Foundations and Trends in Robotics}, 7\penalty0
  (1--2):\penalty0 1--179, 2018.

\bibitem[Pathak et~al.(2018)Pathak, Mahmoudieh, Luo, Agrawal, Chen, Shentu,
  Shelhamer, Malik, Efros, and Darrell]{pathak2018zeroshot}
Pathak, D., Mahmoudieh, P., Luo, G., Agrawal, P., Chen, D., Shentu, Y.,
  Shelhamer, E., Malik, J., Efros, A.~A., and Darrell, T.
\newblock Zero-shot visual imitation.
\newblock In \emph{Proceedings of International Conference on Learning
  Representations (ICLR)}, 2018.

\bibitem[Peng et~al.(2018)Peng, Abbeel, Levine, and van~de
  Panne]{peng2018deepmimic}
Peng, X.~B., Abbeel, P., Levine, S., and van~de Panne, M.
\newblock Deepmimic: Example-guided deep reinforcement learning of
  physics-based character skills.
\newblock \emph{ACM Transactions on Graphics}, 37\penalty0 (4):\penalty0
  143:1--143:14, 2018.

\bibitem[Peng et~al.(2019)Peng, Kanazawa, Toyer, Abbeel, and
  Levine]{peng2019variational}
Peng, X.~B., Kanazawa, A., Toyer, S., Abbeel, P., and Levine, S.
\newblock Variational discriminator bottleneck: Improving imitation learning,
  inverse {RL}, and {GAN}s by constraining information flow.
\newblock In \emph{Proceedings of International Conference on Learning
  Representations (ICLR)}, 2019.

\bibitem[Pomerleau(1989)]{pomerleau1989alvinn}
Pomerleau, D.~A.
\newblock {ALVINN}: An autonomous land vehicle in a neural network.
\newblock In \emph{Proceedings of Neural Information Processing Systems
  (NeurIPS)}, 1989.

\bibitem[Rhinehart et~al.(2020)Rhinehart, McAllister, and
  Levine]{rhinehart2020deep}
Rhinehart, N., McAllister, R., and Levine, S.
\newblock Deep imitative models for flexible inference, planning, and control.
\newblock In \emph{Proceedings of International Conference on Learning
  Representations (ICLR)}, 2020.

\bibitem[Ross et~al.(2011)Ross, Gordon, and Bagnell]{ross2011reduction}
Ross, S., Gordon, G.~J., and Bagnell, J.~A.
\newblock A reduction of imitation learning and structured prediction to
  no-regret online learning.
\newblock In \emph{International Conference on Artificial Intelligence and
  Statistics (AISTATS)}, 2011.

\bibitem[Rybkin et~al.(2019)Rybkin, Pertsch, Derpanis, Daniilidis, and
  Jaegle]{rybkin2019learning}
Rybkin, O., Pertsch, K., Derpanis, K.~G., Daniilidis, K., and Jaegle, A.
\newblock Learning what you can do before doing anything.
\newblock In \emph{Proceedings of International Conference on Learning
  Representations (ICLR)}, 2019.

\bibitem[Schmeckpeper et~al.(2020)Schmeckpeper, Xie, Rybkin, Tian, Daniilidis,
  Levine, and Finn]{schmeckpeper2020learning}
Schmeckpeper, K., Xie, A., Rybkin, O., Tian, S., Daniilidis, K., Levine, S.,
  and Finn, C.
\newblock Learning predictive models from observation and interaction.
\newblock In \emph{Proceedings of European Conference on Computer Vision
  (ECCV)}, 2020.

\bibitem[Sermanet et~al.(2017)Sermanet, Lynch, Chebotar, Hsu, Jang, Schaal, and
  Levine]{sermanet2017time}
Sermanet, P., Lynch, C., Chebotar, Y., Hsu, J., Jang, E., Schaal, S., and
  Levine, S.
\newblock Time-contrastive networks: Self-supervised learning from video.
\newblock In \emph{Proceedings of IEEE International Conference on Robotics and
  Automation}, 2017.

\bibitem[Silver et~al.(2016)Silver, Huang, Maddison, Guez, Sifre, Driessche,
  Schrittwieser, Antonoglou, Panneershelvam, Lanctot, Dieleman, Grewe, Nham,
  Kalchbrenner, Sutskever, Lillicrap, Leach, Kavukcuoglu, Graepel, and
  Hassabis]{silver2016mastering}
Silver, D., Huang, A., Maddison, C.~J., Guez, A., Sifre, L., Driessche, G.
  v.~d., Schrittwieser, J., Antonoglou, I., Panneershelvam, V., Lanctot, M.,
  Dieleman, S., Grewe, D., Nham, J., Kalchbrenner, N., Sutskever, I.,
  Lillicrap, T., Leach, M., Kavukcuoglu, K., Graepel, T., and Hassabis, D.
\newblock Mastering the game of {Go} with deep neural networks and tree search.
\newblock \emph{Nature}, 529\penalty0 (7587):\penalty0 484–--489, 2016.

\bibitem[Stone et~al.(2021)Stone, Ramirez, Konolige, and Jonschkowski]{Stone21}
Stone, A., Ramirez, O., Konolige, K., and Jonschkowski, R.
\newblock The distracting control suite -- a challenging benchmark for
  reinforcement learning from pixels.
\newblock \emph{arXiv preprint 2101.02722}, 2021.

\bibitem[Sun et~al.(2019)Sun, Vemula, Boots, and Bagnell]{sun2019provably}
Sun, W., Vemula, A., Boots, B., and Bagnell, J.~A.
\newblock Provably efficient imitation learning from observation alone.
\newblock In \emph{Proceedings of International Conference on Machine Learning
  (ICML)}, 2019.

\bibitem[Sutton \& Barto(2018)Sutton and Barto]{sutton2018reinforcement}
Sutton, R.~S. and Barto, A.~G.
\newblock \emph{Reinforcement Learning: An Introduction}.
\newblock The MIT Press, second edition, 2018.

\bibitem[Syed et~al.(2008)Syed, Bowling, and Schapire]{syed2008apprenticeship}
Syed, U., Bowling, M., and Schapire, R.~E.
\newblock Apprenticeship learning using linear programming.
\newblock In \emph{Proceedings of International Conference on Machine Learning
  (ICML)}, 2008.

\bibitem[Tassa et~al.(2018)Tassa, Doron, Muldal, Erez, Li, de~Las~Casas,
  Budden, Abdolmaleki, Merel, Lefrancq, Lillicrap, and
  Riedmiller]{tassa2018deepmind}
Tassa, Y., Doron, Y., Muldal, A., Erez, T., Li, Y., de~Las~Casas, D., Budden,
  D., Abdolmaleki, A., Merel, J., Lefrancq, A., Lillicrap, T., and Riedmiller,
  M.
\newblock {DeepMind} control suite.
\newblock \emph{arXiv preprint arXiv:1801.00690}, 2018.

\bibitem[Tomasello(1996)]{tomasello1996doapes}
Tomasello, M.
\newblock Do apes ape?
\newblock In Heyes, C.~M. and Jr., B. G.~G. (eds.), \emph{Social Learning in
  Animals: The Roots of Culture}, chapter~15, pp.\  319--346. Academic Press,
  1996.

\bibitem[Torabi et~al.(2018)Torabi, Warnell, and Stone]{torabi2018behavioral}
Torabi, F., Warnell, G., and Stone, P.
\newblock Behavioral cloning from observation.
\newblock In \emph{Proceedings of International Joint Conference on Artificial
  Intelligence}, 2018.

\bibitem[Torabi et~al.(2019{\natexlab{a}})Torabi, Warnell, and
  Stone]{torabi2019generative}
Torabi, F., Warnell, G., and Stone, P.
\newblock Generative adversarial imitation from observation.
\newblock In \emph{Imitation, Intent, and Interaction (I3) (ICML Workshop)},
  2019{\natexlab{a}}.

\bibitem[Torabi et~al.(2019{\natexlab{b}})Torabi, Warnell, and
  Stone]{torabi2019recent}
Torabi, F., Warnell, G., and Stone, P.
\newblock Recent advances in imitation learning from observation.
\newblock In \emph{Proceedings of International Joint Conference on Artificial
  Intelligence}, 2019{\natexlab{b}}.

\bibitem[Vinyals et~al.(2019)Vinyals, Babuschkin, Czarnecki, Mathieu, Dudzik,
  Chung, Choi, Powell, Ewalds, Georgie, Oh, Horgan, Kroiss, Danihelka, Huang,
  Sifre, Cai, Agapiou, Jaderberg, Vezhnevets, Leblond, Pohlen, Dalibard,
  Budden, Sulsky, Molloy, Paine, Gulcehre, Wang, Pfaff, Wu, Ring, Yogatama,
  W{\"u}nsch, McKinney, Smith, Schaul, Lillicrap, Kavukcuoglu, Hassabis, Apps,
  and Silver]{vinyals2019grandmaster}
Vinyals, O., Babuschkin, I., Czarnecki, W.~M., Mathieu, M., Dudzik, A., Chung,
  J., Choi, D.~H., Powell, R., Ewalds, T., Georgie, P., Oh, J., Horgan, D.,
  Kroiss, M., Danihelka, I., Huang, A., Sifre, L., Cai, T., Agapiou, J.~P.,
  Jaderberg, M., Vezhnevets, A.~S., Leblond, R., Pohlen, T., Dalibard, V.,
  Budden, D., Sulsky, Y., Molloy, J., Paine, T.~L., Gulcehre, C., Wang, Z.,
  Pfaff, T., Wu, Y., Ring, R., Yogatama, D., W{\"u}nsch, D., McKinney, K.,
  Smith, O., Schaul, T., Lillicrap, T., Kavukcuoglu, K., Hassabis, D., Apps,
  C., and Silver, D.
\newblock Grandmaster level in {StarCraft} {II} using multi-agent reinforcement
  learning.
\newblock \emph{Nature}, 575\penalty0 (7782):\penalty0 350--–354, 2019.

\bibitem[Wang et~al.(2017)Wang, Merel, Reed, Wayne, de~Freitas, and
  Heess]{wang2017robust}
Wang, Z., Merel, J., Reed, S., Wayne, G., de~Freitas, N., and Heess, N.
\newblock Robust imitation of diverse behaviors.
\newblock In \emph{Proceedings of Neural Information Processing Systems
  (NeurIPS)}, 2017.

\bibitem[Yu et~al.(2020)Yu, Lyu, and Tsang]{yu2020intrinsic}
Yu, X., Lyu, Y., and Tsang, I.~W.
\newblock Intrinsic reward driven imitation learning via generative model.
\newblock In \emph{Proceedings of International Conference on Machine Learning
  (ICML)}, 2020.

\bibitem[Zhu et~al.(2020)Zhu, Lin, Dai, and Zhou]{zhu2020offpolicy}
Zhu, Z., Lin, K., Dai, B., and Zhou, J.
\newblock Off-policy imitation learning from observations.
\newblock In \emph{Proceedings of Neural Information Processing Systems
  (NeurIPS)}, 2020.

\bibitem[Ziebart(2010)]{ziebart2010modeling}
Ziebart, B.~D.
\newblock \emph{Modeling Purposeful Adaptive Behavior with the Principle of
  Maximum Causal Entropy}.
\newblock PhD thesis, Carnegie Mellon University, 2010.

\bibitem[Ziebart et~al.(2008)Ziebart, Maas, Bagnell, and
  Dey]{ziebart2008maximum}
Ziebart, B.~D., Maas, A., Bagnell, J.~A., and Dey, A.~K.
\newblock Maximum entropy inverse reinforcement learning.
\newblock In \emph{Proceedings of AAAI Conference on Artificial Intelligence},
  2008.

\bibitem[Zolna et~al.(2020)Zolna, Reed, Novikov, Colmenarejo, Budden, Cabi,
  Denil, de~Freitas, and Wang]{zolna2020taskrelevant}
Zolna, K., Reed, S., Novikov, A., Colmenarejo, S.~G., Budden, D., Cabi, S.,
  Denil, M., de~Freitas, N., and Wang, Z.
\newblock Task-relevant adversarial imitation learning.
\newblock In \emph{Conference on Robotic Learning (CoRL)}, 2020.

\end{thebibliography}
\bibliographystyle{icml2021}

\newpage

\onecolumn

\vskip 0.3in

\appendix
\appendixpage

\section{Derivation of the FORM policy gradient}
\label{sec:policy_gradient}

In this section, we derive the following result (given in Sec.~\ref{sec:optimizing_form} of the main text):

\begin{align}
\label{eq:policy_grad_short}
\nabla_\theta \mathcal{J}_\textsc{form}(\pi^I_{\theta}) = \mathbb{E}_{\tau \sim \pi^I_{\theta}} \bigg[ \rho_\textsc{form} \sum_{t \ge 0} \nabla_\theta \log \pi^I_\theta (a_t | x_t) \bigg].
\end{align}

This result is useful because it establishes that the gradient of the imitator policy's parameters does not depend on gradients of either the imitator or demonstrator components of the FORM reward. Because of this, we can safely learn these models in parallel to policy optimization without introducing any bias: this is at the heart of the FORM algorithm. As in the main text, we present our result in terms of models of the form $p(x_t | x_{t-1}, a_{t-1})$ and policies of the form $\pi(a_t | x_t)$, but the results hold without loss of generality to models and policies that depend on states and actions arbitrarily far back into the past.

First, note that for a trajectory $\tau=(x_0, a_0, x_1, a_1, ..., x_{T-1}, a_{T-1})$ produced by a policy with parameters $\theta$:

\begin{align}
\label{eq:policy_identity}
\nabla_\theta p_\theta(\tau) = p_\theta(\tau) \nabla_\theta \sum_{t \ge 0} \log \pi_\theta^I (a_t | x_t).
\end{align}

This result can be shown by decomposing $\tau$ into causal conditional probabilities for state and action (see e.g. \citealt{ziebart2010modeling}):

\begin{align*}
\nabla_\theta p_\theta(\tau) & = \nabla_\theta  \bigg[ \prod_{t \ge 0} p(x_t | x_{t-1}, a_{t-1}) \pi_\theta (a_t | x_t) \bigg] \\
& = \prod_{t \ge 0} p(x_t | x_{t-1}, a_{t-1}) \nabla_\theta \bigg( \prod_{t \ge 0} \pi_\theta (a_t | x_t) \bigg) + \prod_{t \ge 0} \pi_\theta (a_t | x_t) \nabla_\theta \bigg( \prod_{t \ge 0} p(x_t | x_{t-1}, a_{t-1}) \bigg) \\
& = \prod_{t \ge 0} p(x_t | x_{t-1}, a_{t-1}) \nabla_\theta \bigg( \prod_{t \ge 0} \pi_\theta (a_t | x_t) \bigg) \\
& = \frac{p_\theta (\tau)}{\prod_{t \ge 0} \pi_\theta (a_t | x_t)} \nabla_\theta \bigg( \prod_{t \ge 0} \pi_\theta (a_t | x_t) \bigg) \\
& = p_\theta (\tau) \nabla_\theta \log \prod_{t \ge 0} \pi_\theta (a_t | x_t) \\
& = p_\theta (\tau) \nabla_\theta \sum_{t \ge 0} \log \pi_\theta (a_t | x_t).
\end{align*}

With this identity, we can now show the main result. First, we write the policy gradient in terms of the demonstrator and imitator components of the return:

\begin{align}
\label{eq:form_grad}
\nabla_\theta \mathcal{J}(\pi^I_{\theta}) & = \nabla_\theta \mathbb{E}_{\tau \sim \pi_\theta^I} \bigg [ \log p^D(X) - \log p^I_\theta(X) \bigg]
\end{align}

For the first component, the results follows directly from the identity in equation (\ref{eq:policy_identity}) and the log-derivative trick:

\begin{align}
\label{eq:demonstrator_grad}
\nonumber
\nabla_\theta \mathbb{E}_{\tau \sim \pi_\theta^I} \log p^D(X) & = \nabla_\theta \int_{X, A} p_\theta^I(A, X) \log p^D(X) \\
\nonumber
& = \int_{X, A} \nabla_\theta \bigg( p_\theta^I(A, X) \log p^D(X) \bigg) \\
\nonumber
& = \int_{X, A} p_\theta^I(A, X) \nabla_\theta \log p^D(X) + \log p^D(X) \nabla_\theta  p_\theta^I(A, X) \\
\nonumber
& = \int_{X, A} \log p^D(X) \nabla_\theta  p_\theta^I(A, X) \\
\nonumber
& = \int_{X, A} \log p^D(X) p_\theta^I (A, X) \nabla_\theta \sum_{t \ge 0} \log \pi_\theta^I (a_t | x_t) \\
& = \mathbb{E}_{\tau \sim \pi_\theta^I} \bigg[ \log p^D(X) \nabla_\theta \sum_{t \ge 0} \log \pi_\theta^I (a_t | x_t) \bigg]
\end{align}

The derivation of the gradient for the second component follows a similar pattern:

\begin{align*}
\nabla_\theta \mathbb{E}_{\tau \sim \pi_\theta^I} \log p^I_\theta(X) & = \nabla_\theta \int_{X, A} p_\theta^I(A, X) \log p^I_\theta(X) \\
& = \int_{X, A} \nabla_\theta \bigg( p_\theta^I(A, X) \log p^I_\theta(X) \bigg) \\
& = \int_{X, A} p_\theta^I(A, X) \nabla_\theta \bigg( \log p^I_\theta(X) \bigg) + \log p^I_\theta(X) \nabla_\theta \bigg( p_\theta^I(A, X) \bigg) \\
& = \int_{X, A} p_\theta^I(X) \nabla_\theta \bigg( \log p^I_\theta(X) \bigg) + \int_{X, A} \log p^I_\theta(X) \nabla_\theta \bigg( p_\theta^I(A, X) \bigg) \\
\end{align*}

The first of these two integrals vanishes:

\begin{align*}
\int_{X, A} p_\theta^I(X) \nabla_\theta \bigg( \log p^I_\theta(X) \bigg) & = \int_{X} p_\theta^I(X) \nabla_\theta \bigg( \log p^I_\theta(X) \bigg) \\
& = \int_{X} p_\theta^I(X) \frac{\nabla_\theta p^I_\theta(X)}{p^I_\theta(X)} \\
& = \int_{X} \nabla_\theta p^I_\theta(X) \\
& = \nabla_\theta \int_{X} p^I_\theta(X) = \nabla_\theta 1 = 0,
\end{align*}

leaving: 

\begin{align}
\label{eq:imitator_grad}
\int_{X, A} \log p^I_\theta(X) \nabla_\theta \bigg( p_\theta^I(A, X) \bigg) & = \mathbb{E}_{\tau \sim \pi_\theta^I} \bigg[ \log p^I(X) \nabla_\theta \sum_{t \ge 0} \log \pi_\theta^I (a_t | x_t) \bigg],
\end{align}

by our identity. By stitching the expressions in equations (\ref{eq:demonstrator_grad}) and (\ref{eq:imitator_grad}) back into equation (\ref{eq:form_grad}), we get the final result:

\begin{align*}
\nabla_\theta \mathcal{J}(\pi^I_{\theta}) & = \nabla_\theta \mathbb{E}_{\tau \sim \pi_\theta^I} \bigg [ \log p^D(X) - \log p^I_\theta(X) \bigg] \\
& = \mathbb{E}_{\tau \sim \pi_\theta^I} \bigg[ \bigg( \log p^D(X) - \log p^I_\theta(X) \bigg) \nabla_\theta \sum_{t \ge 0} \log \pi_\theta^I (a_t | x_t) \bigg] \\
& = \mathbb{E}_{\tau \sim \pi^I_{\theta}} \bigg[ \rho_\textsc{form} \sum_{t \ge 0} \nabla_\theta \log \pi^I_\theta (a_t | x_t) \bigg].
\end{align*}

\section{Additional results}

\subsection{Additional distractor baselines}
\begin{figure*}[t]
    \centering
    \includegraphics[keepaspectratio,width=1.0\linewidth]{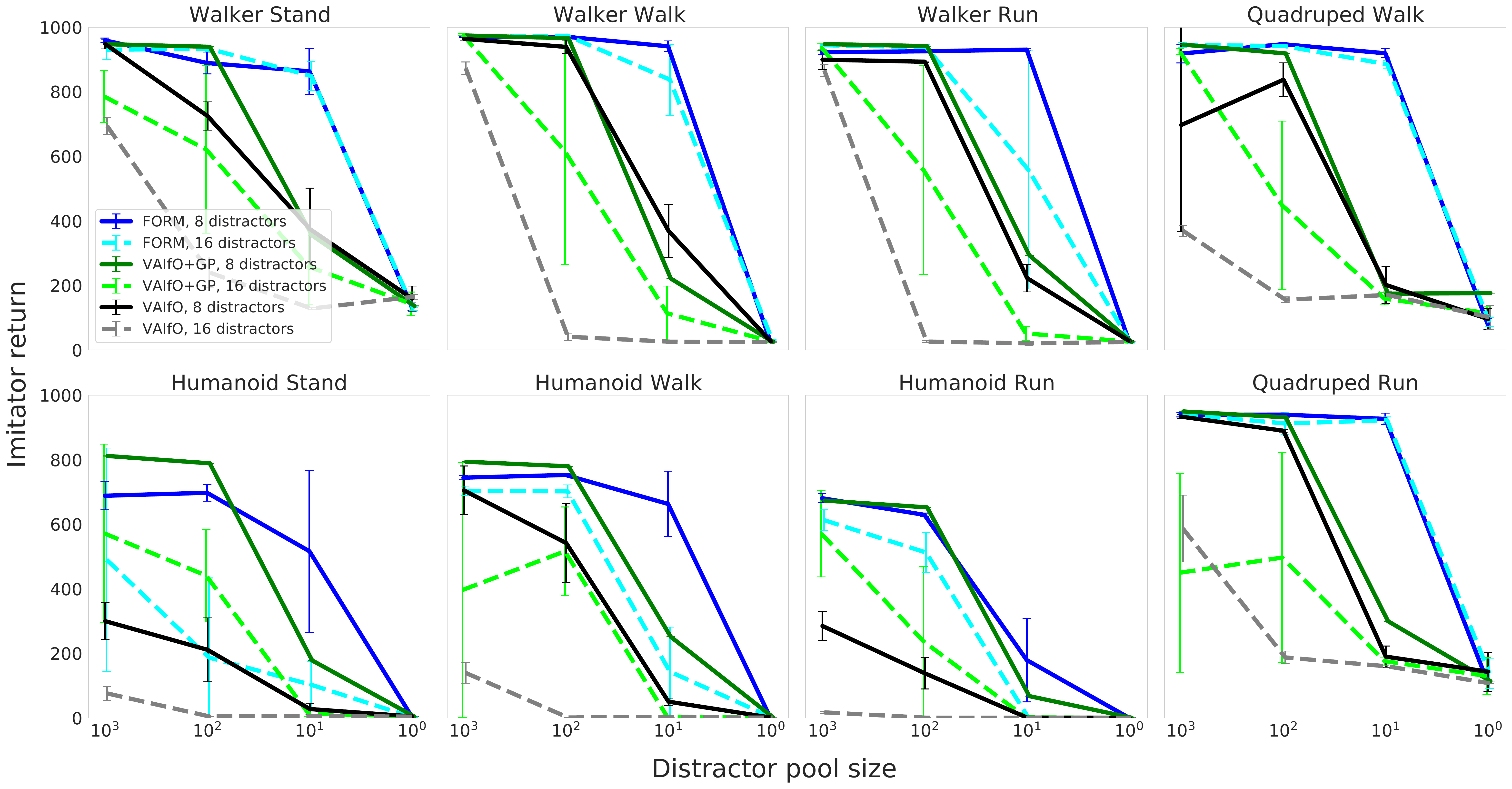}
    \vspace{-8pt}
    \caption{Performance of FORM compared to VAIfO (GAIfO with a variational discriminator bottleneck) and VAIfO+GP in the presence of distractor features.}
    \label{fig:distractors_bc}
\end{figure*}

\begin{figure*}[t]
    \centering
    \includegraphics[keepaspectratio,width=1.0\linewidth]{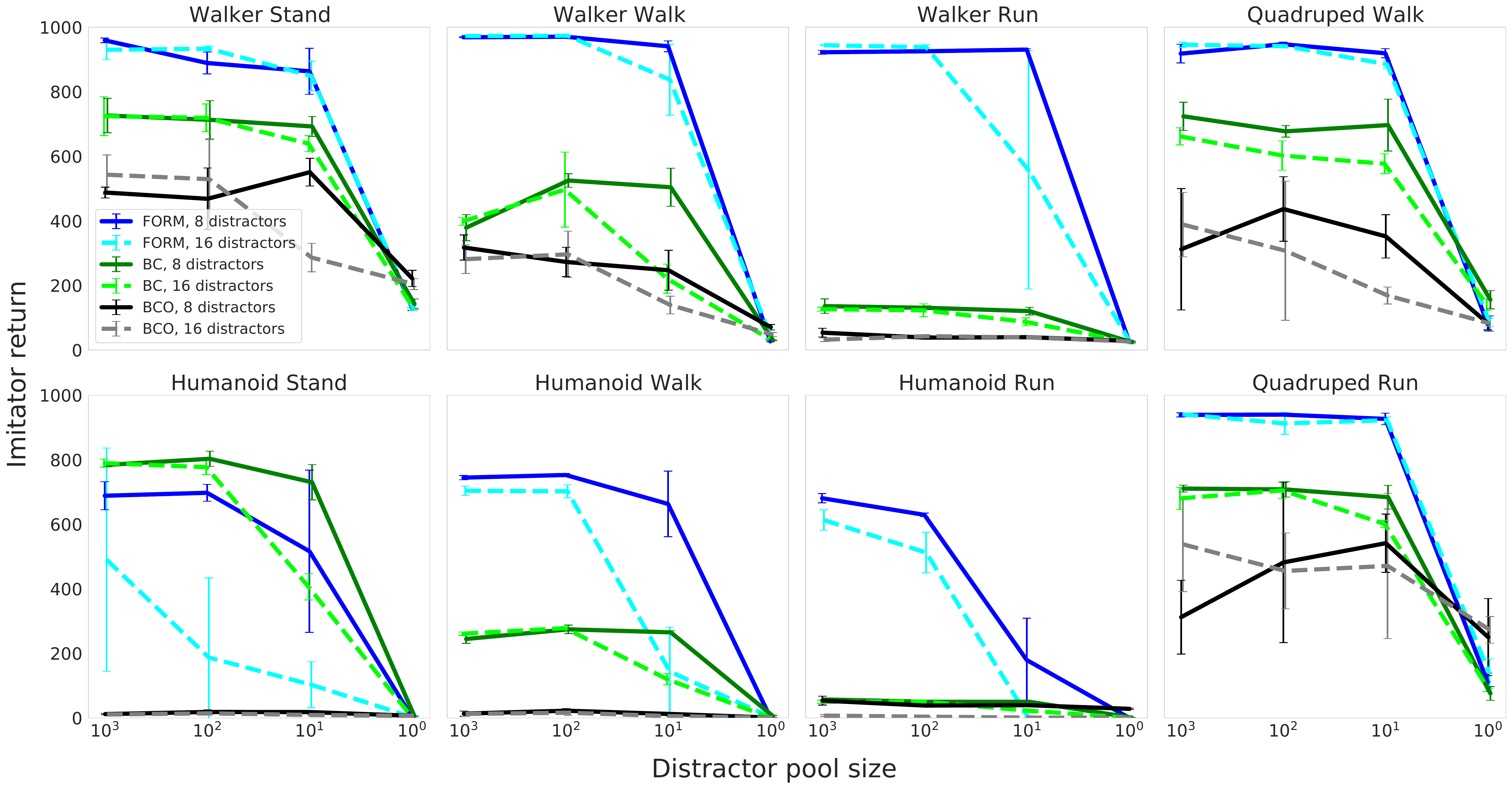}
    \vspace{-8pt}
    \caption{Performance of FORM compared to BC and BCO in the presence of distractor features.}
    \label{fig:distractors_vail}
\end{figure*}

We include comparisons to all baseline methods on the distractor experiments. All experiments are conducted identically to the experiment described in Section~\ref{sec:results} of the main paper. We include FORM in each plot for ease of comparison. Plots depict FORM vs. BC vs. BCO (Fig.~\ref{fig:distractors_bc}) and FORM vs. VAIfO vs. VAIfO+GP (Fig.~\ref{fig:distractors_vail}).

\subsection{Analysis of FORM regularization hyperparameters}
We conducted several additional experiments to estimate the effect of FORM's regularization on the results presented in the main paper, including the effect of autoregressive noise weight on FORM performance (Tab.~\ref{table:autoregressive_ablation}), the effect of maximum overshooting offset on FORM performance (Tab.~\ref{table:overshooting_ablation}), and the effect of L2 regularization (Tab.~\ref{table:l2_ablation}). Each of these contributes a small amount overall, but we found that moderate regularization was important to produce demonstrator models that could effectively guide imitation. We suspect moderate regularization is necessary to prevents the demonstrator log likelihood $\log p^D(x_t | x_{t-1})$ from going to $-\infty$ on transitions not present in the demonstrations. If that were to occur, imitation with a learned model on new data would be very difficult.

\subsection{GAIfO: single vs. two timestep models}
All results presented for GAIfO in the main paper input a single timestep to the discriminator. We made this choice as GAIfO is more susceptible to overfitting when more timesteps are presented as input (for an intuition for why this may happen, see Section~\ref{sec:gail_occupancy} in the main paper). The results of a comparison between GAIfO and GAIfO+GP in the single or two-frame setting are shown in Table~\ref{table:gail_frames}. In early experiments, we observed even more dramatic overfitting when using contexts of length greater than 2.

\section{Experimental Details}

All experiments were conducted in JAX \cite{jax2018github} using tools from the DeepMind JAX ecosystem \cite{deepmind2020jax}. Below we list details of our experimental setup not included in the main paper for reasons of space.

\subsection{Distributed Training}

We train all experiments in a distributed manner: 1 GPU learner updates its parameters from a batch of 64 rollouts pulled from experience replay. Each rollout is 100 timesteps, and the replay buffer stores a maximum of 10,000 rollouts at any time. 50 actors running on CPU execute the environment and push rollouts to the replay buffer. To simplify the implementation, we use the same setup to process offline trajectories (demonstrations) as well as online trajectories (imitator experience). For BC, we keep a single replay buffer and all actors sample a recorded episode and push rollouts to it. For GAIfO and BCO, we keep two replay buffers, one set for the live environment and one set for the demonstration trajectories, and use separate sets of actors to push rollouts to each replay buffer.

\subsection {Network Architectures}

The agents we describe below use a shared architecture to encode observations. The observation encoder:
\begin{itemize}
    \setlength{\itemsep}{0pt}
    \setlength{\parskip}{0pt}
    \setlength{\parsep}{0pt} 
    \item flattens and concatenates its inputs,
    \item linearly projects this observation vector to 256 dimensions,
    \item (optionally) applies layer norm,
    \item activates using a tanh nonlinearity, and
    \item (optionally) further encodes with some number of fully-connected hidden layers of size 256,
    \item activates using an ELU nonlinearity.
\end{itemize}

\subsection{MPO}

We train all policies using MPO \cite{abdolmaleki2018maximum}. This holds for both expert (demonstrator) policies and for the online policy optimization components of both IRL methods (FORM and GAIfO). 

Our MPO agent uses independent policy and critic networks. Both policy and critic networks use an observation encoder with layer norm and one additional hidden layer.

The policy linearly projects the encoded observation to parameterize the mean and scale of a Gaussian action distribution; we ensure that the scale doesn't collapse by transforming it with softplus and adding a minimum value: the scale output is given by $\log (1 + \exp \sigma) + 10^{-4}$.

The critic concatenates the encoded observation with the sampled action (activated with tanh), linearly projects to 256, applies layer norm and tanh again, then further encodes this with a 3-layer MLP with 256-width hidden units ELU activations to produce a (scalar) value output.

To improve stability, we use separate target networks for MPO's policy and critic. We update target networks every 200 gradient updates. MPO uses samples from its Q-function to compare actions at a particular state: we use 20 samples to make each estimate. The critic is then trained using episodic returns computed using Retrace with $\lambda = 1$ and a discount factor of $0.99$. MPO uses independent KL terms to constrain the mean and the scale of the policy: we use constraint weights $\epsilon_{\textrm{mean}} = 0.005$ and $\epsilon_{\textrm{scale}} = 0.00001$ for the two terms, and a shared temperature of $\epsilon_{\textrm{temp}} = 0.1$

We optimize both policy and critic with Adam and a fixed learning rate of $10^{-4}$; we update the temperature and mean/scale duals with Adam and a fixed learning rate of $10^{-3}$.

\subsection{FORM effect model training}

We train two next-step effect models in the FORM agent: one offline on the demonstrations and one online on the live environment and current policy. We found that regularizing the generative models was important to produce good imitation. As described in the main text, we used three simple forms of regularization: (i) L2 regularization, (ii) training on data generated by agent rollouts, i.e. using the network output at a timestep as the input at the next during training , (iii) a form of observation overshooting, i.e. predicting the observations at multiple future timesteps. Rather than overshooting autoregressively, we pass an additional label to the output head, indicating which offset $\delta$ should be predicted. Given input at time $t$, the network is trained to predict the observation at time $t+\delta$. We train the model to predict all offsets in $[1, 5]$. Overshooting is used only during training of the demonstrator and imitator generative models: the reward term reflects the log-likelihood of only the next step.

We also use standardization to ensure the input observations are well-conditioned, as the range of observations varies considerably from task to task on the Control Suite. The use of standardization also prevents imitators from exploiting the structure of the input data to produce misleadingly good results. For example: the observation with largest magnitude on Cheetah Run is the forward velocity, which corresponds almost perfectly to the underlying task reward. An agent that mimics only the forward velocity (while ignoring all other signals) or that maximizes all signals (without imitating) will perform well on the task if the raw observations are used. Results from a strategy like this are misleading in the sense that they may perform well on Cheetah Run because of the design of the observations, not because a general imitation strategy has been learned. Standardization ensures that the mean and standard deviation of all signals are roughly constant, preventing agents from exploiting signals like this. We use standardization for both FORM and GAIL in all experiments in the main paper. We estimate the mean and standard deviation used for standardization by maintain an exponentially decaying running estimate of these quantities for each dimension in the observation (with a decay of $0.99$ per batch of unrolls).

\begin{table}[]
\begin{center}
\begin{tabular}{|l|l|l|l|}
\hline
              & $\beta_{\textrm{ar}}=0.01$ & $\beta_{\textrm{ar}}=0.1$ & $\beta_{\textrm{ar}}=1.0$ \\ \hline
Humanoid run  & 676.02 \err{8.8}           & 668.4 \err{19.4}          & 694.0 \err{10.8}   \\ \hline
Quadruped run & 952.7 \err{3.6}            & 945.4 \err{2.4}           & 955.5 \err{0.4} \\ \hline
\end{tabular}
\caption{Effect of autoregressive noise weight $\beta_{\textrm{ar}}$ on imitator return. }
\label{table:autoregressive_ablation}
\end{center}
\end{table}

\begin{table}[]
\begin{center}
\begin{tabular}{|l|l|l|l|}
\hline
              & $\delta_{\text{max}}=1$ & $\delta_{\text{max}}=3$ & $\delta_{\text{max}}=5$ \\ \hline
Humanoid run  & 266.1 \err{190.5}       & 645.9 \err{32.1}        & 615.8 \err{17.4}  \\ \hline
Quadruped run & 920.57 \err{11.6}       & 920.15 \err{1.2}        & 921.52 \err{9.8} \\ \hline
\end{tabular}
\caption{Effect of maximum overshooting offset ($\delta_{\text{max}}$) on imitator return. All experiments in the paper use $\delta_{\text{max}}=5$.}
\label{table:overshooting_ablation}
\end{center}
\end{table}

\begin{table}[]
\begin{center}
\begin{tabular}{|l|l|l|l|}
\hline
              & $\text{weight}=0.01$ & $\text{weight}=0.1$ & $\text{weight}=1.0$ \\ \hline
Humanoid run  & 682.9 \err{16.8}     & 694.0 \err{10.8}    & 602.5 \err{74.3}  \\ \hline
Quadruped run & 945.4 \err{2.4}      & 953.0 \err{2.8}     & 918.4 \err{0.7} \\ \hline
\end{tabular}
\caption{Effect of L2 regularization weight on imitator return.}
\label{table:l2_ablation}
\end{center}
\end{table}

Including training on model rollouts, overshooting, and standardization, the models maximize $\frac{1}{5} \sum_{\delta=1}^{\delta_{\text{max}}=5} \log p(\sigma(x_{t+\delta}) \mid \sigma(x_t)) + \beta_{\textrm{ar}} \log p(\sigma(x_{t+1}) \mid \sigma(\tilde{x}_t))$, where $\sigma$ is the standardization operator. The first term here is the maximum likelihood, and the second term is the autoregressive regularizer, which conditions on the model's own output $\tilde{x}_t \sim p(\cdot \mid x_{t-1})$. $\beta_{\textrm{ar}}$ was tuned per environment with a grid search over $[0.01, 0.1, 1]$ (see Appendix Table~\ref{table:autoregressive_ablation} for the results of the sweep on two representative domains). We always used overshooting of 5 in the experiments in the paper, as this value generally produced good results in early experiments (see Appendix Table~\ref{table:overshooting_ablation} for ablation results on two representative domains). We found in early experiments that regularization as a whole helped prevent model overfitting (as measured on held-out demonstrator data) and generally led to more stable imitation. 

At each timestep, observations are standardized and encoded as described above. To compute $\log p(\sigma(x_{t+\delta}) \mid \sigma(x_t))$ for a given offset $\delta$, we concatenate a one-hot encoding of $\delta$ to the encoding of $\sigma(x_t)$, then process this with an additional hidden layer of width 256 and linearly project each predicted observation to parameterize a four-component diagonal Gaussian mixture. The scale term of each Gaussian is transformed with softplus to ensure it is non-negative and is added to a small bias term of $10^{-4}$ to avoid degeneracy.

The models are trained with Adam and a fixed learning rate of $10^{-4}$, with an additional $L_2$ regularizer whose coefficient was tuned per-environment by a grid search over $[0.01, 0.1, 1]$ (see Appendix Table~\ref{table:l2_ablation} for the results of the sweep on two representative domains).

\subsection{GAIfO}

\definecolor{Gray}{gray}{0.9}
\begin{table*}[t]
\begin{center}
\begin{tabular}{|l|l|l|l|l|}
\hline
                & GAIfO (1 frame)            & GAIfO (2 frames)           & GAIfO+GP (1 frame)         & GAIfO+GP (2 frames)        \\ \hline \hline
Reacher Easy    & 869.9 \err{48.6}           &  916.1 \err{54.0}          & 915.9 \err{37.8}           & 922.9 \err{15.5}           \\ \hline 
\rowcolor{Gray}
Reacher Hard    & 818.7 \err{11.3}           &  779.0 \err{44.2}          & 783.7 \err{119.7}          & 837.4 \err{23.5}           \\ \hline
Cheetah Run     & 607.6 \err{429.6}          &  5.5 \err{3.4}             & 921.3 \err{6.9}            & 920.4 \err{6.3}            \\ \hline
\rowcolor{Gray}
Quadruped Walk  & 672.6 \err{409.8}          &  125.0 \err{69.0}          & 963.6 \err{4.8}            & 966.2 \err{4.2}            \\ \hline
Quadruped Run   & 952.5 \err{7.5}            &  168.0 \err{25.9}          & 952.3 \err{2.1}            & 951.0 \err{4.6}            \\ \hline
\rowcolor{Gray}
Hopper Stand    & 400.0 \err{164.3}          &  324.3 \err{42.7}          & 748.5 \err{224.1}          & 947.7 \err{7.5}            \\ \hline
Hopper Hop      & 689.2 \err{10.0}           &  683.1 \err{18.6}          & 694.4 \err{0.3}            & 708.9 \err{7.2}            \\ \hline
\rowcolor{Gray}
Walker Stand    & 989.4 \err{1.5}            &  990.3 \err{1.5}           & 985.4 \err{1.6}            & 989.1 \err{0.8}            \\ \hline
Walker Walk     & 976.5 \err{2.8}            &  982.6 \err{0.9}           & 981.6 \err{1.4}            & 977.7 \err{1.0}            \\ \hline
\rowcolor{Gray}
Walker Run      & 949.5 \err{5.6}            & 953.7 \err{1.3}            & 947.6 \err{5.5}            & 945.4 \err{6.2}            \\ \hline
Humanoid Stand  & 4.9   \err{1.0}            & 5.2 \err{0.4}              & 856.2 \err{15.5}           & 697.4 \err{167.8}          \\ \hline
\rowcolor{Gray}
Humanoid Walk   & 1.2   \err{0.4}            & 1.4 \err{0.4}              & 798.4 \err{1.0}            & 792.0 \err{9.4}             \\ \hline
Humanoid Run    & 0.6   \err{0.0}            & 0.7 \err{0.1}              & 683.4 \err{6.9}            & 676.9 \err{17.5}            \\ \hline
\end{tabular}
\caption{GAIfO results when conditioned on one or two frames. We report one frame results in the main table, as this setting was generally stabler and produced the best overall results on the humanoid tasks.}
\vspace{-12pt}
\label{table:gail_frames}
\end{center}
\end{table*}

The discriminator network encodes the current observation (as described above), without layer norm or extra hidden layers, and then applies a two-layer MLP decoder with hidden a width of 256 units to produce the discriminator log odds. As in FORM, we standardize the input observations. We tried training GAIfO using single and two frame input: we generally found better performance using single frames, notably on the humanoid tasks (see Appendix Table~\ref{table:gail_frames}).

GAIfO required regularization to perform adequately on many tasks. We used a gradient penalty on the decoder MLP. The final objective that we maximized for the discriminator was $\log p(\textrm{expert} \mid \tau_{\textrm{expert}}) + \log p(\textrm{imitator} \mid \tau_{\textrm{imitator}}) + \beta_{\textrm{gp}} \lvert \nabla \textrm{decoder}(\textrm{interpolate}(\tau_\textrm{expert}, \tau_\textrm{imitator})) \rvert_2 / 256$ where $\beta_{\textrm{gp}}$ is tuned per environment (typically $10$) and interpolate is a function which mixes the encodings of the expert and imitator observations with random weights sampled each update. We generally observed worse performance when using two frames than one frame without the gradient penalty (Appendix Table~\ref{table:gail_frames}). When using a variational bottleneck, we add an additional hard KL constraint loss term, using a learned weighting term $\alpha$ that is optimized via gradient ascent to keep the bound hard. The value of the constraint itself is set via a hyperparameter $\epsilon$: we swept the value of this constraint in [0, 0.01, 0.1, 1.0, 10.0] and generally obtained best results using a value of 1.0. We report results using $\epsilon=1.0$ throughout.

The discriminator is trained with Adam and a fixed learning rate of $10^{-4}$. Its output is used as the intrinsic reward by the underlying RL agent at each timestep after applying a softplus transform: $\log (1 + p(\textrm{expert} \mid x_t))$.

\subsection{BC and BCO}

In behavioral cloning, we train a Gaussian policy parameterized by a 3-layer MLP. This is the same architecture used for the policy of all other imitation methods. The policy is trained via maximum likelihood to predict the expert actions on trajectories sampled from the recorded demonstrations.

For BCO, we additionally train an inverse model. The inverse model is trained on environment transitions from the learned the policy. The inverse model is then used to predict actions on the expert trajectories, and the policy is updated via the BC objective.

Both the inverse model and policy are updated with Adam \cite{kingma2014adam} and a fixed learning rate of $10^{-4}$.

\section{Gym and Control Suite as Imitation Benchmark Domains}
\label{sec:gym_vs_dmc}

Here, we evaluate methods on the DeepMind Control Suite, but many imitation learning methods are evaluated on the OpenAI Gym Mujoco benchmark \cite{brockman2016gym}. The Control Suite sidesteps two limitations of evaluation on the Gym, which are not always acknowledged in the imitation learning literature, and which make it hard to interpret results.

First, Gym tasks include early termination conditions. For example, Gym's Humanoid task terminates when the agent's head falls below a certain height. Early termination can be helpful for speeding up agent training, but in the context of IRL it introduces a confound: an agent may learn the task by modelling and maximizing the expert's reward function or by making the episode last as long as possible. Any IRL algorithm that produces strictly positive rewards or is otherwise biased to produce longer episodes, can perform well on these tasks while ignoring the expert. GAIL and GAIfO with a softplus discriminator nonlinearity, a typical choice on continuous control domains, fall into this category \cite{kostrikov2019discriminator}. In contrast, episodes on the Control Suite have a fixed duration of 1000 timesteps. 

Second, all Gym domains use very stereotyped initial state distributions: agents are initialized in a single, stable configuration plus a small amount of noise. This means that there is essentially no variation between the configurations seen in the expert demonstrations and on evaluation episodes, which means that even methods that are known to generalize poorly to configurations not seen in the expert data (such as BC \cite{ross2011reduction}) can produce good results on apparently held-out data. In contrast, initial states in the Control Suite are sampled uniformly over the whole configuration space, resulting in a fairly large variation between episodes, especially towards the beginning of episodes. Similar limitations of the Gym control suite environment as a benchmark for control and reinforcement learning are discussed in \cite{mania2018simple}.

\end{document}